\documentclass{article}
\usepackage{ijcai25}

\usepackage{times}
\usepackage{soul}
\usepackage{url}
\usepackage[hidelinks]{hyperref}
\usepackage[utf8]{inputenc}
\usepackage[small]{caption}
\usepackage{graphicx}
\usepackage{amsmath}
\usepackage{amsthm}
\usepackage{booktabs}
\usepackage{algorithm}
\usepackage{algorithmic}
\usepackage[switch]{lineno}

\usepackage{natbib}

\usepackage{subcaption}
\usepackage{multirow}
\usepackage{enumitem}
\usepackage{adjustbox}
\usepackage{xspace}
\usepackage{colortbl}
\usepackage{lipsum}
\usepackage{listings}
\usepackage{xcolor}         
\usepackage{amssymb}
\usepackage{mathtools}
\usepackage{mathabx}
\usepackage{changepage}

\newcommand{\ie}{\textit{i}.\textit{e}., }
\newcommand{\eg}{\textit{e}.\textit{g}., }
\newcommand{\viz}{\textit{viz}., }
\usepackage{pifont}
\newcommand{\cmark}{\ding{51}}%
\newcommand{\sname}{StarFT\xspace}
\definecolor{mygreen}{rgb}{0, 0.6823, 0.7215}
\definecolor{tbd}{rgb}{0.7, 0.11, 0.11}

\usepackage{newfloat}
\usepackage{listings}
\DeclareCaptionStyle{ruled}{labelfont=normalfont,labelsep=colon,strut=off} 
\floatstyle{ruled}
\newfloat{listing}{tb}{lst}{}
\floatname{listing}{Listing}

\renewcommand{\subsubsection}{\vspace{0.05in}\noindent\textbf}

\title{StarFT: Robust Fine-tuning of Zero-shot Models via Spuriosity Alignment}

\author{
Younghyun Kim$^1$\thanks{Equal contribution}{\quad}
Jongheon Jeong$^{2\,*}${\quad}
Sangkyung Kwak$^3$\\
Kyungmin Lee$^4${\quad}
Juho Lee$^4${\quad}
Jinwoo Shin$^4$\\
\affiliations
$^1$Samsung{\quad}
$^2$Korea University{\quad}
$^3$General Robotics{\quad}
$^4$KAIST\\
\emails
yh990220.kim@samsung.com{\quad}
jonghj@korea.ac.kr
\\
sangkyung.kwak@generalrobotics.company{\quad}
\{kyungmnlee, juholee, jinwoos\}@kaist.ac.kr
}

\begin{document}

\maketitle

\begin{abstract}
Learning robust representations from data often requires scale, which has led to the success of recent zero-shot models such as CLIP. However, the obtained robustness can easily be deteriorated when these models are fine-tuned on other downstream tasks (\eg of smaller scales). Previous works often interpret this phenomenon in the context of domain shift, developing fine-tuning methods that aim to preserve the original domain as much as possible. However, in a different context, fine-tuned models with limited data are also prone to learning features that are spurious to humans, such as background or texture. In this paper, we propose \sname (Spurious Textual Alignment Regularization), a novel framework for fine-tuning zero-shot models to enhance robustness by preventing them from \emph{learning spuriosity}. 
We introduce a regularization that aligns the output distribution for spuriosity-injected labels with the original zero-shot model, ensuring that the model is not induced to extract irrelevant features further from these descriptions.
We leverage recent language models to get such spuriosity-injected labels by generating alternative textual descriptions that highlight potentially confounding features.
Extensive experiments validate the robust generalization of \sname and its emerging properties: zero-shot group robustness and improved zero-shot classification. Notably, \sname boosts both worst-group and average accuracy by $14.30\%$ and $3.02\%$, respectively, in the Waterbirds group shift scenario, where other robust fine-tuning baselines show even degraded performance.\footnote{Code is available at \url{https://github.com/alinlab/StarFT}.}
\end{abstract}

\section{Introduction}\label{sec:intro}
Large-scale vision-language models~\citep{radford2021learning, jia2021scaling, zhai2023sigmoid} pre-trained on massive image-caption pairs are shown to have rich representations that generalize to a wide range of tasks, even without fine-tuning on task-specific data (\ie zero-shot generalization).
These zero-shot models, such as CLIP~\citep{radford2021learning}, have demonstrated impressive performance on diverse downstream tasks (defined by a set of textual prompts) without any fine-tuning on a specific target dataset. 
More intriguingly, zero-shot models are further reported to achieve unprecedented robustness across a range of benchmarks involving distribution shifts, which have been a major challenge in the literature~\citep{taori2020measuring, miller2021accuracy}. 
This suggests that the ability to generalize on ``out-of-distribution'' (OOD) inputs may be an emergent property at scale of data.

Although existing zero-shot models provide reasonable performance on diverse tasks, one is often tempted to further \emph{fine-tune} the models in practice when there are task-specific data available to improve their in-distribution (ID) performance.
While such fine-tuning methods effectively enhance ID performance, they are also known to compromise the OOD robustness of the original zero-shot models~\citep{bommasani2022opportunities, wortsman2022robust}. 
{As such, efforts have been recently made to understand the underlying causes of degradation in OOD robustness and mitigate the issue, which is referred to as \emph{robust fine-tuning}.
For example, \citet{goyal2022finetune} have shown that aligning fine-tuning objective with the pre-training stage can improve robustness, and several other works have introduced additional regularization terms~\citep{mao2022contextaware, nushi2018towards}.}

{However, in a broader context, the lack of model robustness is often attributed to learning \emph{spurious features}~\citep{geirhos2020shortcut, jaini2024intriguing, wichmann2023deep}, \ie features that are not aligned with human decision-making but are present in the training data: \eg background~\citep{xiao2021noise}, texture~\citep{geirhos2019imagenet}, and resolution~\citep{touvron2022fixing}.
Existing robust fine-tuning approaches overlook this contributing factor that fine-tuned models depend on confounding decision rules. 
As it is likely that existing zero-shot models and their fine-tuned derivatives also possess certain types of spuriosity that may affect their robustness, further research has been demanded to explore and address them, particularly in the context of broader OOD robustness benchmarks.}
\begin{figure*}[t]
\centering\small
\includegraphics[width=0.95\linewidth]{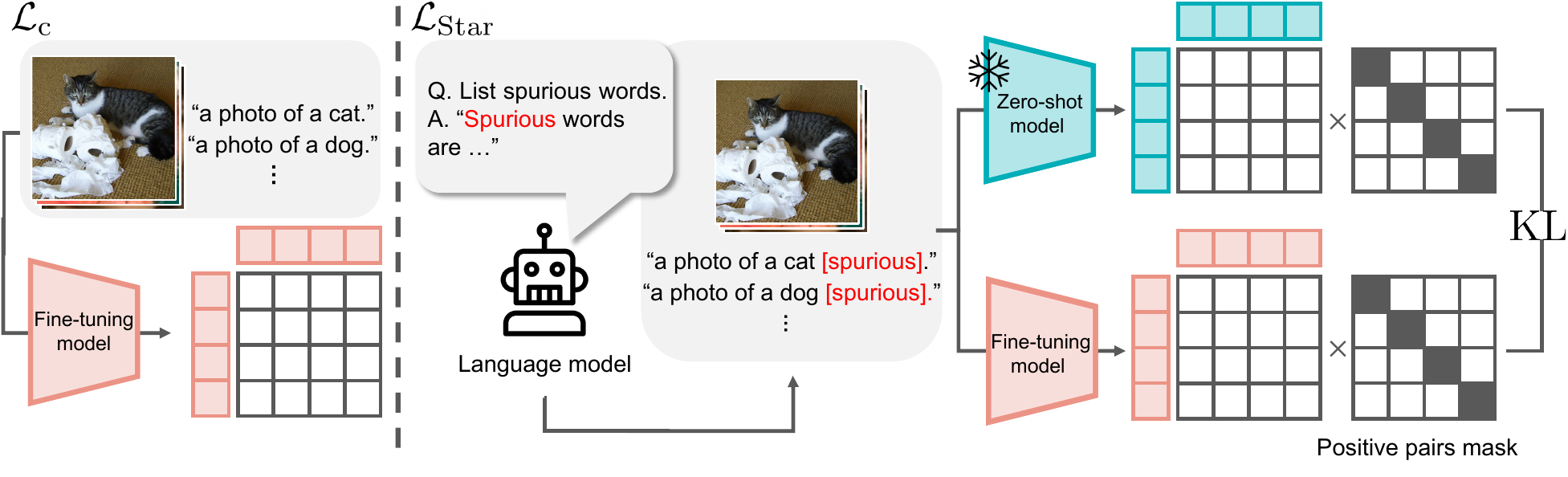}
\caption{
\textbf{Overview of \sname.} Aside from the base contrastive objective $\mathcal{L}_{\textrm{c}}$, we propose a novel spuriosity textual alignment regularization $\mathcal{L}_{\textrm{Star}}$. We first extract spurious textual descriptions from language model, and corrupt the label textual descriptions. We then prevent fine-tuned models from learning spuriosity by minimizing the KL divergence of negative pairs' corrupted textual descriptions.
}
\label{fig:method}
\end{figure*}

\subsubsection{Contribution. }
Motivated by this, we aim to interpret the degradation in OOD robustness of zero-shot models by focusing on the notion of \emph{spuriosity.} 
We guide fine-tuned models to avoid constructing unnecessary decision rules so that the models are more closely aligned with human decision making, which results in improved OOD robustness across diverse benchmarks.
We mitigate the models' reliance on spuriosity by specifying irrelevant textual descriptions that they should not learn during fine-tuning.
We identify the confounding features, such as background, with the aid of language models (LMs).

Specifically, we propose \textit{Spurious Textual Alignment Regularization} (\sname), a novel framework tackling spuriosity for robust fine-tuning of zero-shot models, \eg CLIP.
We generate spurious textual descriptions by querying LMs using only general information, such as ``image classification,'' without any other task-specific prompts, requiring no additional efforts to construct prompts.
Then, we inject the obtained textual spuriosity into the label textual descriptions and regularize fine-tuned models to align the logits distribution of corrupted textual descriptions with zero-shot models.
By directly providing spurious cues to the models and preventing models from extracting such cues during the fine-tuning process, we retrieve the obtained robustness of large-scale pre-trained zero-shot models.

We show that \sname enhances various aspects of zero-shot models: OOD robustness, group shift robustness, zero-shot classification, and transfer learning. Although devising an advanced fine-tuning method for CLIP is a popular research topic recently, 
there exists no such ``universally-good'' method in the literature to the best of our knowledge.

Our contributions are:
\begin{itemize}[leftmargin=5.5mm]
    \item From the observation that fine-tuned models learn spuriosity (Section~\ref{subsec:motivation}), we propose a novel regularization loss that enforces fine-tuned models not to learn the spuriosity (Section~\ref{subsec:our-method}).
    \item We demonstrate that spuriosity textual alignment has indeed improve OOD robustness, as supported by our experiments in domain shift benchmarks (Section~\ref{subsec:domain_shifts}).
    \item We also explore the application of spuriosity textual alignment, achieving zero-shot group robustness (Section~\ref{subsec:zero-shot_robustness});
    \sname demonstrates group robustness in the Waterbirds dataset without having seen any Waterbirds data, where background bias is artificially injected.
    \item Finally, we show that our method enjoys a broader usage by applying it to zero-shot classification (Section~\ref{subsec:zero-shot_classification}) and transfer learning scenarios (Section~\ref{subsec:transfer-learning}).
\end{itemize}

\section{Related Work}\label{sec:related-work}

\subsubsection{Out-of-distribution generalization. }
To deploy machine learning models for real-world applications, the generalization ability to unseen data distribution is crucial~\citep{wiles2022a}. 
To remedy the performance degradation under distribution shifts, extensive efforts have been proposed to enhance the performance under benchmarks that focus on evaluating robustness~\citep{torralba2011unbiased, recht2019imagenet, hendrycks2020augmix, shankar2020evaluating, hendrycks2021faces, tramèr2019adversarial, paul2021vision, mao2022robust, wang2021tent}.
Prior works aim to increase OOD robustness by training with sophisticated data augmentations~\citep{hendrycks2020augmix, hendrycks2021faces}, adversarial training~\citep{tramèr2019adversarial}, using advanced network architecture~\citep{paul2021vision, mao2022robust} or extra information from test-time samples~\citep{wang2021tent}.
Despite the tremendous efforts, there still exists a clear gap between the ID and OOD accuracy~\citep{miller2021accuracy}.
Recent works on zero-shot models~\citep{radford2021learning, jia2021scaling, zhai2023sigmoid} have shown that the existing gap between ID and OOD performance can be notably reduced by scaling up the data curation, demonstrating large improvements in various robustness benchmarks. 
Our work is based upon these advances, by focusing on robust fine-tuning of recent zero-shot models.    

\subsubsection{Robust fine-tuning of zero-shot models. }
Motivated by the fact that fine-tuning zero-shot models is often at the cost of OOD generalization~\citep{andreassen2021evolution, kumar2022finetuning, wortsman2022robust}, various works have explored techniques to preserve the robustness of the zero-shot model while improving its ID accuracy~\citep{li2018explicit, wortsman2022robust, kumar2022finetuning, tian2023trainable, goyal2022finetune, mao2022contextaware, nam2024lipsumft, oh2024calibrated, choi2024autoft}. 
Overall, they have studied post-hoc approaches~\citep{wortsman2022robust, tian2023trainable}, training schemes~\citep{kumar2022finetuning, goyal2022finetune, choi2024autoft}, and regularization schemes~\citep{mao2022contextaware, nam2024lipsumft, oh2024calibrated} to better preserve the zero-shot model as prior knowledge. 
For instances, WiSE-FT~\citep{wortsman2022robust} considers a weight ensembling between the zero-shot and fine-tuned models, and \citet{tian2023trainable} have proposed a projection of the fine-tuned weights to be close to the zero-shot model; both in a post-hoc manner. 
FLYP~\citep{goyal2022finetune} has shown that aligning fine-tuning objective with those of the pre-training stage can improve the OOD performance; 
AutoFT~\citep{choi2024autoft} considers a bi-level optimization to search for a fine-tuning objective based on small OOD validation data; CAR-FT~\citep{mao2022contextaware} regularizes context distributions induced by zero-shot and fine-tuned models, Lipsum-FT~\citep{nam2024lipsumft} reduces energy gap using random texts, and CaRot~\citep{oh2024calibrated} adds self-distillation regularization.

\begin{figure}[t]
\centering
\begin{subfigure}{0.49\linewidth}
\centering
\includegraphics[width=\linewidth]{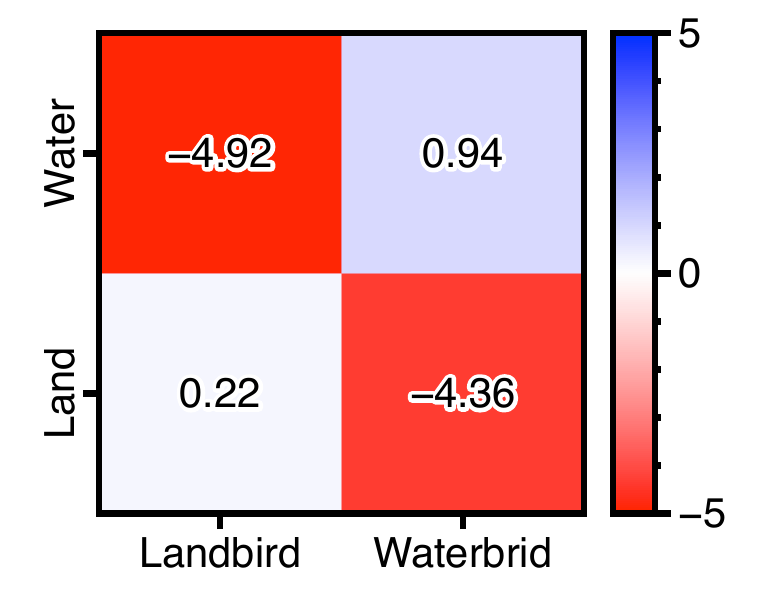}
\caption{Waterbirds}
\end{subfigure}
\begin{subfigure}{0.49\linewidth}
\centering
\includegraphics[width=\linewidth]{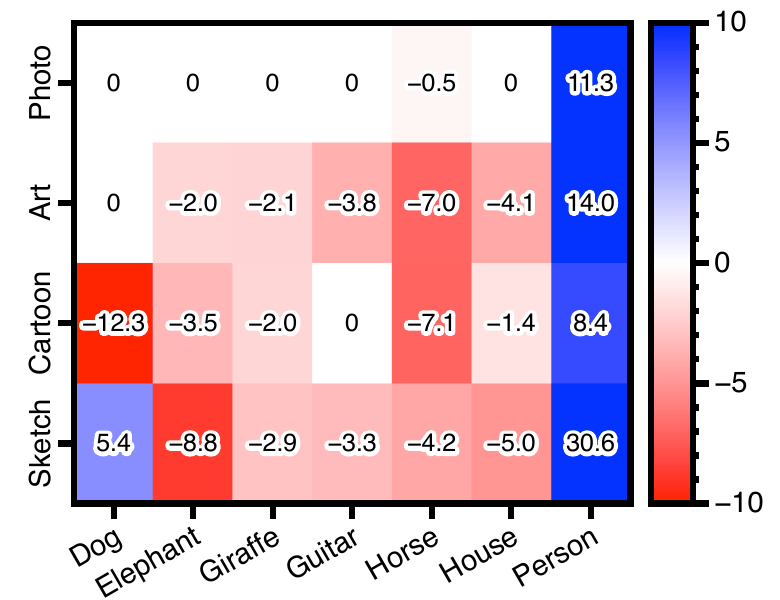}
\caption{PACS}
\end{subfigure}
\caption{\textbf{Subgroup accuracies in group shift benchmarks.} Differences of subgroup accuracies (\%) between zero-shot and FLYP fine-tuned models.}
\label{fig:reasoning}
\end{figure}

\section{Method}
\label{sec:method}
We propose \sname, a robust fine-tuning method leveraging the spurious concept. 
Our motivation stems from the weakness of na\"ive fine-tuning objective in Section~\ref{subsec:motivation} and we explain the proposed method in Section~\ref{subsec:our-method}.
Our goal is to enhance generalization ability of fine-tuned models to OOD domains without compromising ID domain performance.
Given a foundation model and a set of spurious concepts, \sname regularizes models to avoid extracting spurious features other than label information, \ie ``\texttt{[class]},'' to preserve the robustness of the pre-trained foundation model. 

Throughout the paper, we consider an open vocabulary image classification task, where the goal is to map an image $I \in \mathcal{I}$ to a label $y \in \mathcal{Y}$ using image-text aligned vision language models like CLIP~\citep{radford2021learning}.
Given image-label pairs, we design label textual description $T \in \mathcal{T}_y$ of image $I$ using templates such as ``\texttt{a photo of a [class]}'' with class names of each label $y$.

\subsection{Motivation}\label{subsec:motivation}
\subsubsection{Contrastive loss for fine-tuning.}
We start by considering a fine-tuning of the CLIP~\citep{radford2021learning} model on a labeled image dataset, specifically using contrastive loss as done in FLYP~\citep{goyal2022finetune}.
A typical practice for fine-tuning CLIP is to minimize cross-entropy loss; \viz it initializes a new linear head upon the CLIP image embedding to define logits, discarding the textual labels of a given dataset. Therefore, the fine-tuned model is no longer suitable for open vocabulary tasks.
In contrast, employing contrastive loss enables the fine-tuned model to continue processing language inputs as like CLIP.
Formally, given a batch of $N$ image-text pairs $\mathcal{B}=\{(I_i, T_i)\}_{i=1}^N$, let us denote $\boldsymbol{f}_\theta:\mathcal{I}\rightarrow\mathbb{R}^d$ an image encoder and $\boldsymbol{g}_\theta:\mathcal{T}\rightarrow\mathbb{R}^d$ a text encoder.
Then the contrastive loss is given as follows:
\begin{align}
   \mathcal{L}_{\textrm{C}}(\theta;\mathcal{B}) = -\frac{1}{2N}\sum_{i=1}^N\bigg( \log \tfrac{e^{\mathbf{x}_i\cdot \mathbf{y}_i / \tau} }{\sum_{j=1}^N e^{\mathbf{x}_i\cdot \mathbf{y}_j / \tau} }
   + \log \tfrac{e^{\mathbf{x}_i\cdot \mathbf{y}_i / \tau} }{\sum_{j=1}^N e^{\mathbf{x}_j\cdot \mathbf{y}_i / \tau} } \bigg)\text{,} 
\end{align}\label{eq:contrastive}
where $\mathbf{x}_i = \frac{\boldsymbol{f}_\theta(I_i)}{\|\boldsymbol{f}_\theta(I_i)\|_2}$ and $\mathbf{y}_i = \frac{\boldsymbol{g}_\theta(T_i)}{\|\boldsymbol{g}_\theta(T_i)\|_2}$ are $\ell_2$-normalized image and text embeddings, and $\tau>0$ is a temperature. 
When applied for fine-tuning, it first transforms the class labels of the given dataset into texts using some prompt template, \eg ``\texttt{a photo of a [class]},'' and optimizes the contrastive loss \eqref{eq:contrastive} updating both image and text encoders as well as the temperature $\tau$.

However, it is uncertain whether the contrastive fine-tuning loss truly preserves the original ability of CLIP to associate text prompts to images, especially for prompts beyond the template-based prompts derived from the class labels (used during fine-tuning). 
To investigate this, we conduct the following group shift experiment asking whether the fine-tuned model retains its ability to handle diverse textual inputs.

\subsubsection{Spuriosity in fine-tuned zero-shot models.}
We discover that although current robust fine-tuning method achieves higher accuracies in both ID and OOD, it also exhibits shortcut learning that learns spurious correlations from biased data~\citep{sagawa2020distributionally, geirhos2020shortcut} similar to the conventional fine-tuning.
To see this, we examine how the subgroup accuracies of a CLIP model change through its fine-tuning (via FLYP) on ImageNet, as shown in Figure~\ref{fig:reasoning}.
We utilize group shift datasets such as Waterbirds~\citep{sagawa2020distributionally} and PACS~\citep{li2017deeper}, where the samples have extra labels indicating their subgroup domains.
{For example, in Waterbirds, we analyze an ImageNet fine-tuned CLIP using textual prompts such as ``\texttt{a photo of a tench},'' and subsequently test it under group shift scenarios with prompts like ``\texttt{a photo of a waterbird in the mountain}.''
Here, FLYP tends to focus more on confounding features, such as background, rather than the core features, evidenced by its lower performance compared to CLIP on the minor subgroups (\ie landbirds in water background and waterbirds in land background).}

Overall, the results demonstrate that even the most recent advanced robust fine-tuning methods still struggle to avoid spurious correlations during the fine-tuning process. 
This is an unfavorable behavior, especially since these fine-tuned models are often tested in more challenging conditions, such as real-world scenarios~\citep{geirhos2020shortcut}.

\subsection{\sname: Fine-tuning with Spurious Textual Alignment Regularization} 
\label{subsec:our-method}
Motivated by the observation that fine-tuned zero-shot models show shortcut learning, we prevent the models from further constructing unnecessary decision rules during fine-tuning. To achieve this, our method, Spurious Textual Alignment Regularization fine-tuning (\sname), introduces spuriosity suppressing regularization term that restricts model from shortcut learning as illustrated in Figure~\ref{fig:method}.

\subsubsection{Spurious textual alignment regularization.}
To prevent the model from learning spurious features, \ie that leads to learning shortcut, we propose a novel fine-tuning objective named \textit{spurious textual alignment regularization}, which uses the spurious descriptors.
During fine-tuning with contrastive loss, we construct additional captions that include the spurious words and regularize the output of fine-tuning model with the zero-shot model.
In particular, given a batch of $N$ image-text pairs $\mathcal{B}=\{\left(I_i, T_i\right)\}_{i=1}^N$, we construct an additional batch $\mathcal{B}_{\textrm{S}} = \{(I_i, S_i)\}_{i=1}^N$ of image and spuriosity-augmented caption $S_i$, which is generated by attaching spurious keywords at each $T_i$. For instance, given the textual description of class name (\eg ``\texttt{a photo of a [class]}''), we uniformly sample a spurious descriptor and augment to the textual description to generate spuriosity-augmented caption (\eg ``\texttt{a photo of a [class] in the [descriptors]}''). 

Then, we compute regularization loss by measuring the KL divergence between the softmax outputs of fine-tuning and zero-shot models on $\mathcal{B}_S$.  
Specifically, we get the similarities between each image $I_i$ and text $S_i$ using fine-tuning model and zero-shot models.
Then, we mask the logits (both fine-tuning and zero-shot) to exclude the pairs that are of same class labels and define 
$q_i$ to be the softmax probability over the column of masked logits. $q_i$ is given as follows:
\begin{equation}
    [q_i]_j = \frac{e^{\mathbf{x}_i \cdot \mathbf{s}_j / \tau}}{\sum_{c(j') \neq c(i)} e^{\mathbf{x}_i \cdot \mathbf{s}_{j'}/ \tau}}\text{,}
 \end{equation}
where $c(i)$ denotes the class label of $I_i$, $\tau$ is a temperature, and
$\mathbf{x}_i=\frac{\boldsymbol{f}_\theta(I_i)}{\|\boldsymbol{f}_\theta(I_i)\|_2}$, $\mathbf{s}_j = \frac{\boldsymbol{g}_\theta(S_j)}{\|\boldsymbol{g}_\theta(S_j)\|_2}$ are image and text embeddings, respectively. 
By computing $q_i$, we represent the relative likelihood of image $I_i$ that is related to the spurious descriptors $S_j$. 
Here, we mask out logits from the true class to address cases when zero-shot models perform poorly initially, so that distilling their confidence can be unfavorable: \eg as shown in Table~\ref{tab:appx_ablation} in Appendix.
We define $\tilde{q}_i$ be the softmax over the masked logits of zero-shot models in a similar manner, and compute the spurious textual alignment regularization (Star) loss by KL divergence between $q_i$ and $\tilde{q}_i$:
\begin{align}\label{eq:starft-offdiag}
    \mathcal{L}_{\mathrm{Star}}(\theta;\mathcal{B}_{S})
    = \frac{1}{N} \sum_{i=1}^N D_{\textrm{KL}} \left( {{\tilde{q}}}_{i} \,\|\, {q}_{i} \right)\text{.}
\end{align}
Finally, we use linear combination of contrastive fine-tuning loss (\ie Eq.~\eqref{eq:contrastive}) and Star regularization loss (\ie Eq.~\eqref{eq:starft-offdiag}) for fine-tuning, defining the objective of StarFT:
\begin{align}    
\mathcal{L}(\theta; \mathcal{B}, \mathcal{B}_S) = \mathcal{L}_{\textrm{C}}(\theta;\mathcal{B}) + \lambda_{\textrm{Star}} \mathcal{L}_{\textrm{Star}}(\theta; \mathcal{B}_S)\text{,}
\end{align}
where $\lambda_{\textrm{Star}}>0$ is a hyperparamter.\footnote{In our experiments, we use $\lambda_{\textrm{Star}} = 0.5$ by default.}
This hyperparamter $\lambda_{\textrm{Star}}$ is linearly decayed during the course of fine-tuning, which helps balancing the tradeoff between ID and OOD accuracy.

\subsubsection{Obtaining spurious descriptors from LMs.}
As we leverage language supervision in contrastive learning objective, it is natural to use rich semantics of languages in aligning spuriosity.
The advances of large language models have open possibilities of constructing some useful concept banks~\citep{menon2022visual, oikarinen2023labelfree} or extracting task-specific spurious words~\citep{adila2024zeroshot} for classifying images.
Motivated by this, we aim to construct a new set of textual concepts that represents spuriosity, regardless of specific domains; see Table~\ref{tab:examples} for examples.
We obtain textual descriptions of spurious concept that models often rely on while making decisions by querying language models. We prompt the LM with the input:

\begin{quote}
\begin{footnotesize}
\begin{verbatim}
Question: List possible spurious 
correlations while classifying natural 
images. Answer in a word.

Answer: 1.
-
\end{verbatim}
\end{footnotesize}
\end{quote}
Given minimal task description, LMs provide spurious concepts like ``background,'' ``texture,'' and ``resolution'' that corresponds to our belief. Then, to directly corrupt the label textual description $T$, we ask LMs for more fine-grained spurious descriptors 
$s$.
For example, for spurious concept ``\texttt{[background],}'' we get spurious descriptors such as ``in the mountains'' or ``on the beach.''
We compose the spurious descriptors set $\mathcal{S}$ for each spurious concept, which is used to corrupt the textual description of an image.

\setlength{\tabcolsep}{0.07in}
\begin{table}[t]
\centering
\begin{tabular}{l|lll}
    \texttt{[background]} & mountains & beach & desert \\ \midrule
    \texttt{[texture]} & rough & smooth & soft \\ \midrule
    \texttt{[resolution]} & blurred & bright & overexposed
\end{tabular}
\caption{{\textbf{Examples of spurious words.} Samples of the obtained spurious words for each spurious concepts. The list of concrete textual descriptions can be found in Appendix~\ref{appx:spurious_list}.}}
\label{tab:examples}
\end{table}

\setlength{\tabcolsep}{0.07in}
\begin{table*}[t]
\centering\small
\begin{tabular}{lc|cccc|c|c|cccc|c}
\toprule
 &  \multicolumn{6}{c|}{ViT-B/16}& \multicolumn{6}{c}{ViT-L/14} \\ \cmidrule(l){2-13} 
Method   & \multicolumn{1}{c|}{IN}    & IN-R & IN-A  & IN-S  & \multicolumn{1}{c|}{IN-V2} & Avg.  & \multicolumn{1}{c|}{IN} & IN-R & IN-A & IN-S & \multicolumn{1}{c|}{IN-V2} & Avg. \\ \midrule
Zeroshot  & 68.3 & \textbf{77.7} & 50.0 & 48.3  & 61.9 & 59.5 & \multicolumn{1}{c|}{75.6} & 87.9 & 70.8 & 59.6 & \multicolumn{1}{c|}{69.9} & 72.0\\
FT        & 81.3  & 71.3 & 44.5 & 49.1 & 71.7 & 59.1 & \multicolumn{1}{c|}{84.7} & 75.4 & 55.7 & 54.4 & \multicolumn{1}{c|}{75.3} & 65.2\\
FLYP      & 82.6  & 71.4 & 48.5 & 49.8 & 72.7 & 60.6 & \multicolumn{1}{c|}{86.2} & 83.8 & 68.9 & 60.2 & \multicolumn{1}{c|}{78.2} & 72.8 \\
CAR-FT    & 81.9 & 75.6 & 50.0 & 51.5  & 72.8 & 62.5 & \multicolumn{1}{c|}{86.3} & 84.2 & 66.6 & 60.0 & \multicolumn{1}{c|}{76.8} & 71.9 \\
CaRot &
  \textbf{83.1} &
  \underline{76.2} &
  \underline{51.3} &
  \underline{51.9} &
    \textbf{74.3} &
  \underline{63.7}  & \multicolumn{1}{c|}{\textbf{87.0}} & \underline{88.0} & \underline{72.7} & \underline{62.7} & \multicolumn{1}{c|}{\textbf{79.3}} & \underline{75.6}
   \\ \midrule
\rowcolor{mygreen! 10}
\textbf{\sname~(Ours)} &
  \underline{82.9} & 
  \textbf{77.7} &
  \textbf{53.7} &
  \textbf{52.5} &
    \underline{73.8} &
  \textbf{64.4} & \multicolumn{1}{c|}{\underline{86.4}} & \textbf{88.7} & \textbf{73.8} & \textbf{63.2} & \multicolumn{1}{c|}{\underline{78.9}} & \textbf{76.1 }
   \\ \bottomrule
\end{tabular}%
\caption{
\textbf{Evaluation on domain shifts.} 
We report Top-1 accuracies (\%) on ImageNet (IN) and OOD datasets (IN-R, IN-A, IN-S, IN-V2), with their average values (Avg.) for two architectures (ViT-B/16 and ViT-L/14).
\sname outperforms the baselines under domain shifts scenarios on ImageNet. For example, on ViT-B/16, \sname outperforms full fine-tuning by 5.3\% on OOD and 1.6\% on IN. 
We \textbf{bold} and \underline{underline} the top two values in each column.}
\label{tab:domain_shift}
\end{table*}

\begin{table}[t]
\centering\small
\adjustbox{width=\linewidth}{
\begin{tabular}{lcc|cc|cc}\toprule
 & \multicolumn{2}{c|}{Waterbirds} & \multicolumn{2}{c|}{PACS} & \multicolumn{2}{c}{CIFAR-10.02} \\ 
\cmidrule(l){2-3}
\cmidrule(l){4-5}
\cmidrule(l){6-7}
Method   & WG    & Avg.    & WG     & Avg.      & WG     & Avg.  \\ \midrule
Zeroshot & 25.9 & 87.1 & 87.4 & 93.0  &  47.0 & 87.0   \\
FT       & \underline{27.1} & 85.6 & 87.6 & 91.6 &  62.5 & 85.7  \\
FLYP      & 21.5 & 87.2  & 86.0 & 92.5 &  61.0 & 88.2 \\
CAR-FT    & 24.8 & 86.6  & 87.2 & 93.2&  \underline{63.5} & 87.3  \\
CaRot     & \underline{27.1} & \underline{89.5} & \underline{88.5} &  \underline{94.2} &  63.0 & \underline{89.3} \\ \midrule \rowcolor{mygreen! 10}
{\bf \sname~(Ours)}    & \textbf{40.2}   & \textbf{90.1} & \textbf{89.1} & \textbf{94.7} &  \textbf{64.5} & \textbf{90.2}  \\ \bottomrule
\end{tabular}%
}
\caption{
\textbf{Evaluation on group shifts.} 
We report worst-group (WG) and average (Avg.) accuracies (\%) of ImageNet fine-tuned models on Waterbirds, PACS, and CIFAR-10.02 datasets. 
Notably, \sname (Ours) consistently outperforms baselines including pre-trained CLIP without seeing any data from the group shift benchmarks. We \textbf{bold} and \underline{underline} the top two values in each column.
}
\label{tab:group_shift}
\end{table}

\setlength{\tabcolsep}{0.05in}
\begin{table}[t]
\centering\small
\begin{tabular}{lcccc|c}
\toprule
Method   & C-10  & C-100      & Cal101     & \multicolumn{1}{c|}{STL10} & Avg.                 \\ \midrule
Zeroshot  & 90.8   & \underline{68.2}       & \underline{89.6}          & 98.3                & \underline{86.7}   \\
FT        & 87.7   & 63.6    & 85.7          &  95.3                  &83.1  \\
FLYP      & 90.0   & 64.2       & 87.4          & 98.5               &85.0  \\
CAR-FT    & 89.7   & 65.9     & 88.2         & 96.7                 &85.2\\
CaRot     & \underline{91.1}    & 66.7      & 89.0         & \underline{98.7}                &86.5 \\ \midrule\rowcolor{mygreen! 10}
{\bf \sname (Ours)}      & \textbf{91.4} &\textbf{69.0} & \textbf{89.7} & \textbf{99.0} & \textbf{87.3}  \\ \bottomrule
\end{tabular}%
\caption{
\textbf{Zero-shot evaluation.} We evaluate ImageNet fine-tuned models in different zero-shot scenarios on CIFAR-10 (C-10), CIFAR-100 (C-100), Caltech101 (Cal101), and STL10. Notably, without seen any data from the zero-shot benchmarks, ours consistently outperforms baselines including pre-trained CLIP. We \textbf{bold} and \underline{underline} the top two values in each column.
}
\label{tab:zero_shot}
\end{table}

\subsubsection{Comparison with other methods.}
Recent works tackle the robust fine-tuning by introducing additional regularization term, such as CAR-FT~\citep{mao2022contextaware} and CaRot~\citep{nushi2018towards}.
CAR-FT optimizes the cross-entropy loss
with additional dataset-specific context regularization to guide the fine-tuned model towards the zero-shot model.
On the other hand, we regularize spuriosity rather than context, making it scalable to any other datasets.
By directly addressing spuriosity, StarFT achieves better robustness than CAR-FT.
Also, StarFT integrates with contrastive loss by injecting irrelevant information into the label textual descriptions. This direct adaptation eliminates the computational overhead associated with CAR-FT, which relies on averaged weights of context prompts across classes.
CaRot, another recent baseline, optimizes the contrastive loss with additional regularization to maintain the image and textual logit distributions of zero-shot models.
Additionally, CaRot updates zero-shot models using an exponential moving average (EMA).
Due to this EMA update, CaRot fails on benchmarks where zero-shot models suffer.
However, ours does not use EMA update to boost performance, and thus is not affected by whether zero-shot models perform poorly or not.

\section{Experiments}
\label{sec:experiments}
First, we demonstrate the robustness of \sname in diverse distribution shift scenarios (Section~\ref{subsec:domain_shifts}). 
Then, we present that the model fine-tuned with \sname is robust to group shift benchmarks (Section~\ref{subsec:zero-shot_robustness}) and outperforms on various object classification datasets (Section~\ref{subsec:zero-shot_classification}), even outperforming pre-trained zero-shot models.
Lastly, we apply \sname for standard transfer learning tasks (Section~\ref{subsec:transfer-learning}) demonstrating our method's efficacy in various scenarios.

\begin{table*}[t]
\centering\small
\begin{tabular}{lcccccc|c}
\toprule
Method     & Caltech101 & Cars  & Flowers & ImageNet & iWILD & \multicolumn{1}{c|}{FMoW} & Avg. Rank\\ \midrule
Zeroshot   & 87.7     & 64.4 & 81.2  & 68.3    & 8.70 & 20.4 & 6.0\\
FT        & 97.0     & 84.2 & 92.4  & 81.3    & 45.2 & \textbf{68.6} & 3.8\\
FLYP      & \underline{97.1}     & 89.0 & \underline{97.1}  & 82.6    & \underline{48.5} & \textbf{68.6} &\underline{2.2} \\
CAR-FT     & 96.1     & 84.3 & 94.5  & 81.9    & 45.8 & \underline{68.4} &3.7\\
CaRot      & 96.0     & \textbf{89.8}  & 95.7 & \textbf{83.1}    & 40.6 & 51.9&3.3\\ \midrule\rowcolor{mygreen! 10}
{\bf \sname (Ours)}&  \textbf{97.2} & \underline{89.5} & \textbf{97.5}  &  \underline{82.8}   & \textbf{50.1} & \underline{68.4} & \textbf{1.7}\\ \bottomrule 
\end{tabular}%
\caption{
\textbf{Transfer learning.} We evaluate our proposed approach on 6 different transfer learning datasets. We fine-tune with the downstream datasets and report the ID test accuracy. We \textbf{bold} and \underline{underline} the top two values in each column.
}
\label{tab:transfer_learning}
\end{table*}

\subsubsection{Baselines.}
We compare \sname with various fine-tuning methods for pre-trained zero-shot models; standard fine-tuning with classification loss (FT), and robust fine-tuning approaches such as FLYP~\citep{goyal2022finetune}, CAR-FT~\citep{mao2022contextaware}, and CaRot~\citep{oh2024calibrated}.

\subsubsection{Implementation details.}
Throughout experiments, we utilize CLIP~\citep{radford2021learning} ViT-B$/$16 and ViT-L$/$14 trained on the LAION dataset~\citep{schuhmann2021laion} and fine-tune the model using AdamW~\citep{kingma2017adam} optimizer with a cosine learning rate scheduler. 
We train models with a batch size of 512 for ImageNet, while all other datasets use a batch size of 256.
OOD datasets are only used for evaluation, where we select the best-performing model based on ID validation set accuracy.
Across all datasets, we use the same text-templates as CLIP~\citep{radford2021learning} and WiSE-FT~\citep{wortsman2022robust}. 
Further implementation details are in Appendix~\ref{appx:experimental_detail}.

\subsection{Evaluation on domain shifts} \label{subsec:domain_shifts}
\subsubsection{Datasets.} To assess the performance of our approach across domain shifts, we train \sname on ImageNet (IN)~\citep{russakovsky2015imagenet}, which comprises over a million natural images of 1,000 classes.
We then evaluate our fine-tuned models on 4 well-known ImageNet OOD benchmarks: ImageNet-R (IN-R)~\citep{hendrycks2021faces}, ImageNet-A (IN-A)~\citep{hendrycks2021natural}, ImageNet-Sketch (IN-S)~\citep{wang2019learning}, and ImageNetV2 (IN-V2)~\citep{recht2019imagenet}. 
ImageNet-R contains visual renditions such as ``cartoons'' of ImageNet classes while ImageNet-Sketch contains sketches of ImageNet classes.
ImageNet-A consists of naturally occurring samples that are misclassified by ResNet models.
ImageNetV2 is a newly curated test set of ImageNet.
Each of these variants reflects the domain shift of ImageNet.

\subsubsection{Results.}
As shown in Table~\ref{tab:domain_shift}, \sname shows the best OOD average accuracy while maintaining high ID accuracy for both small and large CLIP models.
This highlights the effectiveness of our spurious alignment loss in enhancing domain robustness, providing empirical evidence that eliminating spuriosity during fine-tuning enhances model robustness against distribution shifts.
On ImageNet, with ViT-B$/$16, our method surpasses full fine-tuning (FT) by 5.3\% in OOD average accuracy and by 1.57\% in ID accuracy. 
Similarly, with the larger ViT-L$/$14 architecture, StarFT achieves performance improvements on both ID and OOD datasets.
In fact, ours shows the best OOD average accuracy even when compared to baselines with weight ensembling, \ie via WiSE-FT \citep{wortsman2022robust}, as shown in Table~\ref{tab:ensemble} in Appendix.

\subsection{Evaluation on group shifts}\label{subsec:zero-shot_robustness}
\subsubsection{Datasets.} 
We further evaluate our ImageNet fine-tuned models on three different group shift benchmarks: Waterbirds~\citep{sagawa2020distributionally}, PACS~\citep{li2017deeper}, and CIFAR-10.02~\citep{zhang2022contrastive}. 
As we are given the class labels from these datasets, we construct textual descriptions to perform zero-shot evaluations.
Waterbirds classifies bird images into landbird and waterbird, with each class has two groups based on the background: bird on land background, and bird on water background.
PACS classifies images into 7 categories and each image is from either arts, cartoons, photos or sketches, indicating different groups.
CIFAR-10.02 classifies images into 10 categories.
We follow \citet{zhang2022contrastive} to construct CIFAR-10.02, which combines the original CIFAR-10~\citep{krizhevsky2009learning} and CIFAR-10.2~\citep{lu2020harder} from different data sources, defining each data source as a distinct group.

\subsubsection{Results.}
To verify the effect of our spurious alignment loss in group shifts, which are closely related to spurious correlations, we leverage popular group shift benchmarks as a means of measuring spuriosity.
As our objective is to enhance a wide range of robustness by targeting spuriosity, we do not additionally fine-tune with datasets in group shift benchmarks.
Since all fine-tuned models are capable of zero-shot classification, we assess the spuriosity inherent in fine-tuned models using zero-shot performance on group shift benchmarks.
We report both the {worst group accuracy} (WG) and the average accuracy (Avg.), as the worst group accuracy reflects the robustness of the model across different groups within the data.
Notably, our spurious alignment loss indeed improves the group robustness of the fine-tuned model, resulting in the best WG accuracy among all baselines as depicted in Table~\ref{tab:group_shift}.
It is well known that improving worst group accuracy often comes at the cost of average accuracy~\citep{sagawa2020distributionally}.
However, as opposed to this common drawback of eliminating spuriosity, \sname improves both in the worst group and the average accuracy.
This stands out in the Waterbirds benchmark, where ours narrows the gap between the worst group and the average accuracy to 49.88\%, which is 61.19\% in zero-shot models.

\subsection{Zero-shot classification}\label{subsec:zero-shot_classification}
We also conduct evaluation on zero-shot classification from the ImageNet fine-tuned models on 4 natural image benchmarks: CIFAR-10~\citep{krizhevsky2009learning}, CIFAR-100~\citep{krizhevsky2009learning}, Caltech101~\citep{li_andreeto_ranzato_perona_2022}, and STL10~\citep{coates2011stl}.
Overall, we observe that \sname preserves the generalization ability of zero-shot models even after its fine-tuning.
As shown in Table~\ref{tab:zero_shot}, \sname outperforms the baselines including the base zero-shot models, across all datasets considered.
In CIFAR-100 and Caltech101, although all fine-tuned baselines show deteriorated zero-shot performance compared to the zero-shot models, ours could maintain or even improve their accuracy.

\subsection{Transfer learning}\label{subsec:transfer-learning}
We compare the transferability of \sname between various fine-tuning methods. We use 6 common object classification datasets:
Caltech101~\citep{li_andreeto_ranzato_perona_2022}, StanfordCars~\citep{krause2013stanfordcars}, Flowers102~\citep{nilsback2008automated}, ImageNet~\citep{russakovsky2015imagenet}, WILDS-iWILDCam~\citep{beery2020iwildcam, koh2021wilds}, and WILDS-FMoW~\citep{christie2018functional, koh2021wilds}. 
{In our experiments, we use the same set of spurious descriptors across datasets.}
Table~\ref{tab:transfer_learning} shows the results.
Compared to a strong baseline such as CaRot~\citep{oh2024calibrated}, \sname (Ours) shows slight degradation on Cars and ImageNet datasets.
This is partly due to the CaRot's EMA update on zero-shot models which iteratively self-distills the knowledge of zero-shot models.
However, this approach has drawbacks in scenarios where the zero-shot model performs poorly, such as in WILDS-iWILDCam and WILDS-FMoW.
In these datasets, the zero-shot model struggles with ID accuracies of 8.70\% and 18.7\%, respectively, unlike other datasets.
Consequently, CaRot performs significantly worse than other baselines, while \sname shows consistent improvement across all 6 datasets with the best average rank in transfer learning since ours does not rely on EMA updates.

\subsection{Ablation study}
\label{sec:result}

\subsubsection{Corruption to label textual descriptions.}
We show the effect of each component in our proposed method in Table~\ref{tab:ablation}.
We start from adding regularization to FLYP~\citep{goyal2022finetune} which is the standard contrastive learning objective.
Regularizing clean label textual descriptions without any spurious corruptions improves OOD accuracy but degrades the ID performance as it continues to interfere with learning to fit the in-domain dataset.
Then we append the random textual tokens to label textual descriptions to see the effect of spurious descriptions in improvements. 
Results indicate that adding random tokens in the suffix improves both ID and OOD performance, where adding spurious textual descriptions further boosts the performance.

\begin{table}[t]
\centering\small
\setlength\tabcolsep{2.5pt} 
\adjustbox{width=\linewidth}{
\begin{tabular}{ccccc|ccccc}
\toprule
      &            &      &          & \multicolumn{6}{c}{ImageNet}                                             \\ \cmidrule(l){5-10} 
$\mathcal{L}_{\text{Star}}$ & Suffix     & Mask & Decay & \multicolumn{1}{c|}{IN}   & IN-R & IN-S          & IN-A & \multicolumn{1}{c|}{IN-V2} & Avg. \\ \midrule
-- &   --      &  --    &   --       &  82.6  & 71.4 & 48.5 & 49.8 & \multicolumn{1}{c|}{72.7} & 60.6 \\
 \cmark&       --     &  --    &     --     & 80.4  & 77.3          & 51.5 & 52.4  & \multicolumn{1}{c|}{71.6}  & 63.2 \\
 \cmark     & Random   &  --    &    --      & 82.3   & 77.6          & 52.4 & 52.1 & \multicolumn{1}{c|}{73.2}  & 63.8 \\ \midrule
 \cmark     & Spurious &    --  &  --        & 82.7   & 78.0& 52.8 &52.5  & \multicolumn{1}{c|}{73.5} & 64.2 \\ 
 \cmark     & Spurious &\cmark   &     --     & 82.3  & 78.2          & 54.2 & 52.6 & \multicolumn{1}{c|}{73.3}   & 64.6 \\
 \cmark     & Spurious &   --   &\cmark     &   82.9    &          77.4     &   53.2   & 52.3  &   \multicolumn{1}{c|}{73.8}    &  64.2  \\\rowcolor{mygreen! 10}
 \cmark& Spurious &\cmark&\cmark& 82.9 & 77.7 & 53.7 & 52.5 & \multicolumn{1}{c|}{73.8}  & 64.4 \\ \bottomrule
\end{tabular}%
}
\caption{
\textbf{Ablation on different components.} 
We ablate on different parts of \sname by starting with the baseline, where we regularize the clean label descriptions, and then add random and spurious textual descriptions to corrupt them.
Next, we fix to the spurious suffix and examine the effects of other components: masking positive pairs and diminishing regularization ratio.}
\label{tab:ablation}
\end{table}
\begin{table}[t]
\centering \small
\adjustbox{width=\linewidth}{
\begin{tabular}{cccccccc}
\toprule
           & \multicolumn{6}{c}{ImageNet}                                                             \\ \cmidrule(l){2-7}
Concept    & \multicolumn{1}{c|}{IN}    & IN-R & IN-S  & IN-A  & \multicolumn{1}{c|}{IN-V2} & Avg.  \\ \midrule
\texttt{[background]} & \multicolumn{1}{c|}{82.9}  & 77.7 & 53.7
& {52.5} & \multicolumn{1}{c|}{73.8} & 64.4 \\
\texttt{[texture]}    & \multicolumn{1}{c|}{82.8}  & 78.3 & 53.7 & {53.0} & \multicolumn{1}{c|}{73.8} & 64.7 \\
\texttt{[resolution]} & \multicolumn{1}{c|}{82.7}  & 78.3 & 53.6 & {53.0} & \multicolumn{1}{c|}{73.8} & 64.7 \\ \midrule
all        & \multicolumn{1}{c|}{82.4}  & 78.6 & 53.9 & {53.1} & \multicolumn{1}{c|}{73.5} & 64.8 \\ \bottomrule
\end{tabular}%
}
\caption{\textbf{Ablation on different spurious concepts.}
We ablate on different spurious concepts, including ``background,'' ``texture,'' ``resolution,'' and ``all.''
Adding spurious textual descriptions consistently improves both ID and OOD accuracies.
}
\label{tab:concepts}
\end{table}

\subsubsection{Component-wise analysis.}
We further test two additional components of StarFT in Table~\ref{tab:ablation}: masking and decaying regularization.
For the masking, we observe an overall gain in OOD accuracies on ImageNet, and even larger gains on WILDS-iWILDCam (see Table~\ref{tab:appx_ablation} in Appendix); where there exist many over-confident ``wrong'' positive pairs (as zero-shot model suffers).
Since positive pairs in the mini-batch have very higher confidences than those of negative pairs, it suppresses the model from learning useful knowledge that lies in spurious descriptions, thereby resulting in deterioration of both ID and OOD accuracies.
Regarding the decaying regularization; 
we adopt the decaying $\lambda_{\text{Star}}$ in StarFT primarily due to its effectiveness in preserving ID accuracy. Note that a key practical objective of robust fine-tuning is not only to improve OOD robustness, but also to ensure that ID performance is not compromised. As observed in Table~\ref{tab:ablation}, we find that the gain in ID performance (\textit{e.g.}, $+0.6\%$ ImageNet accuracy) from the decaying strategy often outweighs the slight drop in OOD robustness (\textit{e.g.}, $-0.2\%$ average OOD accuracy).
For an additional ablation study, \eg on the effect of $\lambda$, and spuriosity concepts, see Appendix~\ref{appx:additional_experiment}.

\begin{figure}[t]
\centering\small
\includegraphics[width=\linewidth]{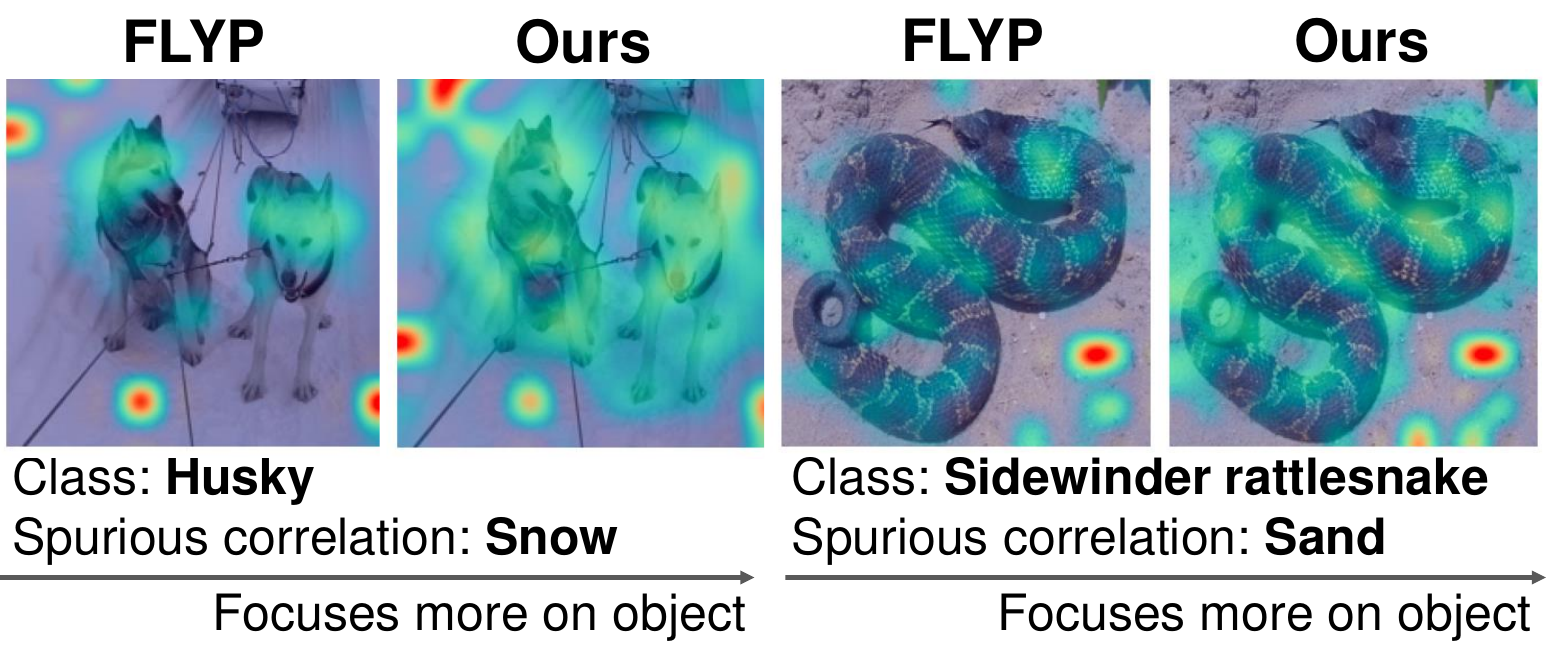}
\caption{
\textbf{Mitigation of spuriosity in ImageNet.} We display the GradCAM~\citep{Selvaraju_2019} of fine-tuned models for comparison. 
Each class has the spurious correlations with background such as ``snow'' in ``husky'' and ``sand'' in ``rattle snake.''
Rather than focusing on mostly background like FLYP, \sname focuses on object itself to make decisions.
}
\label{fig:hook}
\end{figure}

\subsubsection{Choice of different spurious concept.}
By prompting language models, we attain the spurious concept descriptors of ``background,'' ``texture,'' and ``resolution.''
We fine-tune \sname using each of the spurious concepts and observe that adding spurious descriptions consistently improves both the ID and OOD accuracies.
Combination of all concepts results in the slight improvement in OOD accuracy, however, at the cost of ID accuracy, meaning restricting spuriosity too much would sacrifice the ID accuracy.
To study the efficacy of our methods in diverse downstream tasks, we fix our spurious concept to ``background'' throughout the experiments, which shows the best validation ID accuracy.
However, we believe that further investigation on combinations of different spurious concepts would be promising research directions.

\subsubsection{Spurious correlations in ImageNet.} 
We study the spurious correlations of fine-tuned models in ImageNet.
We observe that the \sname mitigates the spuriosity present in the zero-shot models while classifying ImageNet.
To identify the spurious correlations in the zero-shot models, we adopt the setup of \citet{kim2024discovering} to each of ImageNet classes.
We discover that ``snow'' and ``sand'' is spuriously correlated to ``husky'' and  ``rattle snake,'' respectively, indicating a background bias in the zero-shot models.
However, this reliance on the background diminishes in \sname as shown in Figure~\ref{fig:hook}.
Compared to ours, the FLYP does not focus on object than background when classifying items.

\section{Conclusion}
\label{sec:conclusion}
It remains unclear which parts of the knowledge encoded in zero-shot models commit to their effective robustness, and how to preserve it during fine-tuning.
We believe our approach of tackling \emph{spuriosity} during fine-tuning suggests a novel view on understanding the robustness of zero-shot models. 
By devising a textual regularization that aims to prevent models from adopting spurious decision rules, aided by the textual interface of zero-shot models, we could improve the consistency of robustness across many benchmarks, where current methods have been inconsistent.
We open the potential of studying the effect of spuriosity in zero-shot model's fine-tuning on wider notions of robustness. 
We further discuss on Limitations and Broader impacts in Appendix~\ref{appx:limitations}. 

\section*{Acknowledgements}
This work was supported in part by Institute of Information \& communications Technology Planning \& Evaluation (IITP) grant funded by the Korea government(MSIT) (No.RS-2022-II220184, 2022-0-00184, Development and Study of AI Technologies to Inexpensively Conform to Evolving Policy on Ethics) and in part by Center for Applied Research in Artificial Intelligence(CARAI) grant funded by Defense Acquisition Program Administration(DAPA) and Agency for Defense Development(ADD) (UD230017TD). 
Jongheon Jeong acknowledges support from IITP grants funded by the Korea government (MSIT)
(RS-2019-II190079, Artificial Intelligence Graduate School Program (Korea University); 
IITP-2025-RS-2024-00436857, Information Technology Research Center (ITRC);
IITP-2025-RS-2025-02304828, Artificial Intelligence Star Fellowship Support Program to Nurture the Best Talents) and
the Korea Creative Content Agency grant funded by the Ministry of Culture, Sports and Tourism (RS-2024-00345025).

\bibliographystyle{unsrtnat}
\bibliography{references}

\begin{thebibliography}{57}
\providecommand{\natexlab}[1]{#1}
\providecommand{\url}[1]{\texttt{#1}}
\expandafter\ifx\csname urlstyle\endcsname\relax
  \providecommand{\doi}[1]{doi: #1}\else
  \providecommand{\doi}{doi: \begingroup \urlstyle{rm}\Url}\fi

\bibitem[Radford et~al.(2021)Radford, Kim, Hallacy, et~al.]{radford2021learning}
Alec Radford, Jong~Wook Kim, Chris Hallacy, et~al.
\newblock Learning transferable visual models from natural language supervision.
\newblock In \emph{ICML}, 2021.

\bibitem[Jia et~al.(2021)Jia, Yang, Xia, Chen, et~al.]{jia2021scaling}
Chao Jia, Yinfei Yang, Ye~Xia, Yi-Ting Chen, et~al.
\newblock Scaling up visual and vision-language representation learning with noisy text supervision.
\newblock In \emph{International Conference on Machine Learning}, 2021.

\bibitem[Zhai et~al.(2023)Zhai, Mustafa, Kolesnikov, and Beyer]{zhai2023sigmoid}
Xiaohua Zhai, Basil Mustafa, Alexander Kolesnikov, and Lucas Beyer.
\newblock Sigmoid loss for language image pre-training.
\newblock In \emph{Proceedings of the IEEE/CVF International Conference on Computer Vision}, pages 11975--11986, 2023.

\bibitem[Taori et~al.(2020)Taori, Dave, Shankar, Carlini, Recht, and Schmidt]{taori2020measuring}
Rohan Taori, Achal Dave, Vaishaal Shankar, Nicholas Carlini, Benjamin Recht, and Ludwig Schmidt.
\newblock Measuring robustness to natural distribution shifts in image classification.
\newblock In \emph{Neural Information Processing Systems}, 2020.

\bibitem[Miller et~al.(2021)Miller, Taori, Raghunathan, Sagawa, Koh, et~al.]{miller2021accuracy}
John Miller, Rohan Taori, Aditi Raghunathan, Shiori Sagawa, Pang~Wei Koh, et~al.
\newblock Accuracy on the line: On the strong correlation between out-of-distribution and in-distribution generalization.
\newblock In \emph{International Conference on Machine Learning}, 2021.

\bibitem[Bommasani et~al.(2022)Bommasani, Hudson, Adeli, et~al.]{bommasani2022opportunities}
Rishi Bommasani, Drew~A. Hudson, Ehsan Adeli, et~al.
\newblock On the opportunities and risks of foundation models, 2022.

\bibitem[Wortsman et~al.(2022)Wortsman, Ilharco, Kim, Li, et~al.]{wortsman2022robust}
Mitchell Wortsman, Gabriel Ilharco, Jong~Wook Kim, Mike Li, et~al.
\newblock Robust fine-tuning of zero-shot models.
\newblock In \emph{Conference on Computer Vision and Pattern Recognition}, 2022.

\bibitem[Goyal et~al.(2023)Goyal, Kumar, Garg, Kolter, and Raghunathan]{goyal2022finetune}
Sachin Goyal, Ananya Kumar, Sankalp Garg, Zico Kolter, and Aditi Raghunathan.
\newblock Finetune like you pretrain: Improved finetuning of zero-shot vision models.
\newblock In \emph{Conference on Computer Vision and Pattern Recognition}, 2023.

\bibitem[Mao et~al.(2022{\natexlab{a}})Mao, Chen, Jia, Zhang, Xue, and Li]{mao2022contextaware}
Xiaofeng Mao, Yuefeng Chen, Xiaojun Jia, Rong Zhang, Hui Xue, and Zhao Li.
\newblock Context-aware robust fine-tuning.
\newblock \emph{International Journal of Computer Vision}, 2022{\natexlab{a}}.

\bibitem[Nushi et~al.(2018)Nushi, Kamar, and Horvitz]{nushi2018towards}
Besmira Nushi, Ece Kamar, and Eric Horvitz.
\newblock Towards accountable ai: Hybrid human-machine analyses for characterizing system failure.
\newblock In \emph{AAAI Conference on Human Computation and Crowdsourcing}, 2018.

\bibitem[Geirhos et~al.(2020)Geirhos, Jacobsen, Michaelis, et~al.]{geirhos2020shortcut}
Robert Geirhos, J{\"o}rn-Henrik Jacobsen, Claudio Michaelis, et~al.
\newblock Shortcut learning in deep neural networks.
\newblock \emph{Nature Machine Intelligence}, 2\penalty0 (11):\penalty0 665--673, 2020.

\bibitem[Jaini et~al.(2024)Jaini, Clark, and Geirhos]{jaini2024intriguing}
Priyank Jaini, Kevin Clark, and Robert Geirhos.
\newblock Intriguing properties of generative classifiers.
\newblock In \emph{International Conference on Learning Representations}, 2024.

\bibitem[Wichmann and Geirhos(2023)]{wichmann2023deep}
Felix~A. Wichmann and Robert Geirhos.
\newblock Are deep neural networks adequate behavioral models of human visual perception?
\newblock \emph{Annual Review of Vision Science}, pages 501--524, 2023.

\bibitem[Xiao et~al.(2021)Xiao, Engstrom, Ilyas, and Madry]{xiao2021noise}
Kai Xiao, Logan Engstrom, Andrew Ilyas, and Aleksander Madry.
\newblock Noise or signal: The role of image backgrounds in object recognition.
\newblock In \emph{International Conference on Learning Representations}, 2021.

\bibitem[Geirhos et~al.(2019)Geirhos, Rubisch, Michaelis, et~al.]{geirhos2019imagenet}
Robert Geirhos, Patricia Rubisch, Claudio Michaelis, et~al.
\newblock Imagenet-trained cnns are biased towards texture; increasing shape bias improves accuracy and robustness.
\newblock In \emph{International Conference on Learning Representations}, 2019.

\bibitem[Touvron et~al.(2019)Touvron, Vedaldi, Douze, and Jégou]{touvron2022fixing}
Hugo Touvron, Andrea Vedaldi, Matthijs Douze, and Hervé Jégou.
\newblock Fixing the train-test resolution discrepancy.
\newblock In \emph{Neural Information Processing Systems}, 2019.

\bibitem[Wiles et~al.(2022)Wiles, Gowal, Stimberg, Rebuffi, Ktena, Dvijotham, and Cemgil]{wiles2022a}
Olivia Wiles, Sven Gowal, Florian Stimberg, Sylvestre-Alvise Rebuffi, Ira Ktena, Krishnamurthy~Dj Dvijotham, and Ali~Taylan Cemgil.
\newblock A fine-grained analysis on distribution shift.
\newblock In \emph{International Conference on Machine Learning}, 2022.

\bibitem[Torralba and Efros(2011)]{torralba2011unbiased}
Antonio Torralba and Alexei~A. Efros.
\newblock Unbiased look at dataset bias.
\newblock In \emph{Conference on Computer Vision and Pattern Recognition}, 2011.

\bibitem[Recht et~al.(2019)Recht, Roelofs, Schmidt, and Shankar]{recht2019imagenet}
Benjamin Recht, Rebecca Roelofs, Ludwig Schmidt, and Vaishaal Shankar.
\newblock Do imagenet classifiers generalize to imagenet?
\newblock In \emph{International Conference on Machine Learning}, 2019.

\bibitem[Hendrycks et~al.(2020)Hendrycks, Mu, Cubuk, et~al.]{hendrycks2020augmix}
Dan Hendrycks, Norman Mu, Ekin~D. Cubuk, et~al.
\newblock Augmix: A simple data processing method to improve robustness and uncertainty.
\newblock In \emph{International Conference on Learning Representations}, 2020.

\bibitem[Shankar et~al.(2020)Shankar, Roelofs, Mania, Fang, Recht, and Schmidt]{shankar2020evaluating}
Vaishaal Shankar, Rebecca Roelofs, Horia Mania, Alex Fang, Benjamin Recht, and Ludwig Schmidt.
\newblock Evaluating machine accuracy on {I}mage{N}et.
\newblock In \emph{International Conference on Machine Learning}, 2020.

\bibitem[Hendrycks et~al.(2021{\natexlab{a}})Hendrycks, Basart, Mu, et~al.]{hendrycks2021faces}
Dan Hendrycks, Steven Basart, Norman Mu, et~al.
\newblock The many faces of robustness: A critical analysis of out-of-distribution generalization.
\newblock In \emph{International Conference on Computer Vision}, 2021{\natexlab{a}}.

\bibitem[Tramèr and Boneh(2019)]{tramèr2019adversarial}
Florian Tramèr and Dan Boneh.
\newblock Adversarial training and robustness for multiple perturbations.
\newblock In \emph{NeurIPS}, 2019.

\bibitem[Paul and Chen(2021)]{paul2021vision}
Sayak Paul and Pin-Yu Chen.
\newblock Vision transformers are robust learners.
\newblock In \emph{AAAI}, 2021.

\bibitem[Mao et~al.(2022{\natexlab{b}})Mao, Qi, Chen, Li, Duan, Ye, He, and Xue]{mao2022robust}
Xiaofeng Mao, Gege Qi, Yuefeng Chen, Xiaodan Li, Ranjie Duan, Shaokai Ye, Yuan He, and Hui Xue.
\newblock Towards robust vision transformer.
\newblock In \emph{Neural Information Processing Systems}, 2022{\natexlab{b}}.

\bibitem[Wang et~al.(2021)Wang, Shelhamer, Liu, Olshausen, and Darrell]{wang2021tent}
Dequan Wang, Evan Shelhamer, Shaoteng Liu, Bruno Olshausen, and Trevor Darrell.
\newblock Tent: Fully test-time adaptation by entropy minimization.
\newblock In \emph{International Conference on Learning Representations}, 2021.

\bibitem[Andreassen et~al.(2021)Andreassen, Bahri, Neyshabur, and Roelofs]{andreassen2021evolution}
Anders Andreassen, Yasaman Bahri, Behnam Neyshabur, and Rebecca Roelofs.
\newblock The evolution of out-of-distribution robustness throughout fine-tuning, 2021.

\bibitem[Kumar et~al.(2022)Kumar, Raghunathan, Jones, Ma, and Liang]{kumar2022finetuning}
Ananya Kumar, Aditi Raghunathan, Robbie Jones, Tengyu Ma, and Percy Liang.
\newblock Fine-tuning can distort pretrained features and underperform out-of-distribution.
\newblock In \emph{International Conference on Learning Representations}, 2022.

\bibitem[Li et~al.(2018)Li, Grandvalet, and Davoine]{li2018explicit}
Xuhong Li, Yves Grandvalet, and Franck Davoine.
\newblock Explicit inductive bias for transfer learning with convolutional networks.
\newblock In \emph{ICML}, 2018.

\bibitem[Tian et~al.(2023)Tian, Dai, Ma, He, Liu, and Kira]{tian2023trainable}
Junjiao Tian, Xiaoliang Dai, Chih-Yao Ma, Zecheng He, Yen-Cheng Liu, and Zsolt Kira.
\newblock Trainable projected gradient method for robust fine-tuning.
\newblock In \emph{Conference on Computer Vision and Pattern Recognition}, 2023.

\bibitem[Nam et~al.(2024)Nam, Heo, and Lee]{nam2024lipsumft}
Giung Nam, Byeongho Heo, and Juho Lee.
\newblock Lipsum-ft: Robust fine-tuning of zero-shot models using random text guidance.
\newblock In \emph{International Conference on Learning Representations}, 2024.

\bibitem[Oh et~al.(2024)Oh, Lim, Kim, Choo, Hauptmann, Cheng, and Song]{oh2024calibrated}
Changdae Oh, Hyesu Lim, Mijoo Kim, Jaegul Choo, Alexander Hauptmann, Zhi-Qi Cheng, and Kyungwoo Song.
\newblock Towards calibrated robust fine-tuning of vision-language models, 2024.

\bibitem[Choi et~al.(2024)Choi, Lee, Chen, Zhou, Raghunathan, and Finn]{choi2024autoft}
Caroline Choi, Yoonho Lee, Annie Chen, Allan Zhou, Aditi Raghunathan, and Chelsea Finn.
\newblock Autoft: Learning an objective for robust fine-tuning, 2024.

\bibitem[Sagawa et~al.(2020)Sagawa, Koh, Hashimoto, and Liang]{sagawa2020distributionally}
Shiori Sagawa, Pang~Wei Koh, Tatsunori~B Hashimoto, and Percy Liang.
\newblock Distributionally robust neural networks for group shifts: On the importance of regularization for worst-case generalization.
\newblock In \emph{ICLR}, 2020.

\bibitem[Li et~al.(2017)Li, Yang, Song, and Hospedales]{li2017deeper}
Da~Li, Yongxin Yang, Yi-Zhe Song, and Timothy~M. Hospedales.
\newblock Deeper, broader and artier domain generalization.
\newblock In \emph{ICCV}, 2017.

\bibitem[Menon and Vondrick(2023)]{menon2022visual}
Sachit Menon and Carl Vondrick.
\newblock Visual classification via description from large language models.
\newblock In \emph{International Conference on Learning Representations}, 2023.

\bibitem[Oikarinen et~al.(2023)Oikarinen, Das, Nguyen, and Weng]{oikarinen2023labelfree}
Tuomas Oikarinen, Subhro Das, Lam~M. Nguyen, and Tsui-Wei Weng.
\newblock Label-free concept bottleneck models.
\newblock In \emph{International Conference on Learning Representations}, 2023.

\bibitem[Adila et~al.(2024)Adila, Shin, Cai, and Sala]{adila2024zeroshot}
Dyah Adila, Changho Shin, Linrong Cai, and Frederic Sala.
\newblock Zero-shot robustification of zero-shot models.
\newblock In \emph{International Conference on Learning Representations}, 2024.

\bibitem[Schuhmann et~al.(2021)Schuhmann, Vencu, Beaumont, et~al.]{schuhmann2021laion}
Christoph Schuhmann, Richard Vencu, Romain Beaumont, et~al.
\newblock L{AION}-400m: Open dataset of clip-filtered 400 million image-text pairs, 2021.

\bibitem[Kingma and Ba(2017)]{kingma2017adam}
Diederik~P. Kingma and Jimmy Ba.
\newblock Adam: A method for stochastic optimization, 2017.

\bibitem[Russakovsky et~al.(2015)Russakovsky, Deng, Su, et~al.]{russakovsky2015imagenet}
Olga Russakovsky, Jia Deng, Hao Su, et~al.
\newblock Imagenet large scale visual recognition challenge.
\newblock \emph{International Journal of Computer Vision}, 2015.

\bibitem[Hendrycks et~al.(2021{\natexlab{b}})Hendrycks, Zhao, Basart, Steinhardt, and Song]{hendrycks2021natural}
Dan Hendrycks, Kevin Zhao, Steven Basart, Jacob Steinhardt, and Dawn Song.
\newblock Natural adversarial examples.
\newblock In \emph{CVPR}, 2021{\natexlab{b}}.

\bibitem[Wang et~al.(2019)Wang, Ge, Xing, and Lipton]{wang2019learning}
Haohan Wang, Songwei Ge, Eric~P. Xing, and Zachary~C. Lipton.
\newblock Learning robust global representations by penalizing local predictive power.
\newblock In \emph{Neural Information Processing Systems}, 2019.

\bibitem[Zhang and R{\'e}(2022)]{zhang2022contrastive}
Michael Zhang and Christopher R{\'e}.
\newblock Contrastive adapters for foundation model group robustness.
\newblock In \emph{Neural Information Processing Systems}, 2022.

\bibitem[Krizhevsky(2009)]{krizhevsky2009learning}
Alex Krizhevsky.
\newblock Learning multiple layers of features from tiny images, 2009.

\bibitem[Lu et~al.(2020)Lu, Nott, Olson, et~al.]{lu2020harder}
Shangyun Lu, Bradley Nott, Aaron Olson, et~al.
\newblock Harder or different? a closer look at distribution shift in dataset reproduction.
\newblock In \emph{ICML Workshop on Uncertainty and Robustness in Deep Learning}, 2020.

\bibitem[Li et~al.(2022)Li, Andreeto, Ranzato, and Perona]{li_andreeto_ranzato_perona_2022}
Fei-Fei Li, Marco Andreeto, Marc'Aurelio Ranzato, and Pietro Perona.
\newblock Caltech 101, 2022.

\bibitem[Coates et~al.(2011)Coates, Ng, and Lee]{coates2011stl}
Adam Coates, Andrew Ng, and Honglak Lee.
\newblock An analysis of single-layer networks in unsupervised feature learning.
\newblock In \emph{International Conference on Machine Learning}, 2011.

\bibitem[Krause et~al.(2013)Krause, Stark, Deng, and Fei-Fei]{krause2013stanfordcars}
Jonathan Krause, Michael Stark, Jia Deng, and Li~Fei-Fei.
\newblock 3d object representations for fine-grained categorization.
\newblock In \emph{IEEE International Conference on Computer Vision Workshops}, pages 554--561, 2013.

\bibitem[Nilsback and Zisserman(2008)]{nilsback2008automated}
Maria-Elena Nilsback and Andrew Zisserman.
\newblock Automated flower classification over a large number of classes.
\newblock In \emph{2008 Sixth Indian conference on computer vision, graphics \& image processing}, pages 722--729. IEEE, 2008.

\bibitem[Beery et~al.(2020)Beery, Cole, and Gjoka]{beery2020iwildcam}
Sara Beery, Elijah Cole, and Arvi Gjoka.
\newblock The i{WildCam} 2020 competition dataset, 2020.

\bibitem[Koh et~al.(2021)Koh, Sagawa, Marklund, et~al.]{koh2021wilds}
Pang~Wei Koh, Shiori Sagawa, Henrik Marklund, et~al.
\newblock Wilds: A benchmark of in-the-wild distribution shifts.
\newblock In \emph{International Conference on Machine Learning}, 2021.

\bibitem[Christie et~al.(2018)Christie, Fendley, Wilson, and Mukherjee]{christie2018functional}
Gordon Christie, Neil Fendley, James Wilson, and Ryan Mukherjee.
\newblock Functional map of the world.
\newblock In \emph{Conference on Computer Vision and Pattern Recognition}, 2018.

\bibitem[Selvaraju et~al.(2017)Selvaraju, Cogswell, Das, Vedantam, Parikh, and Batra]{Selvaraju_2019}
Ramprasaath~R. Selvaraju, Michael Cogswell, Abhishek Das, Ramakrishna Vedantam, Devi Parikh, and Dhruv Batra.
\newblock Grad-{CAM}: Visual explanations from deep networks via gradient-based localization.
\newblock In \emph{ICCV}, 2017.

\bibitem[Kim et~al.(2024)Kim, Mo, Kim, Lee, Lee, and Shin]{kim2024discovering}
Younghyun Kim, Sangwoo Mo, Minkyu Kim, Kyungmin Lee, Jaeho Lee, and Jinwoo Shin.
\newblock Discovering and mitigating visual biases through keyword explanation.
\newblock In \emph{Conference on Computer Vision and Pattern Recognition}, 2024.

\bibitem[Grattafiori et~al.(2024)Grattafiori, Dubey, Jauhri, Pandey, et~al.]{grattafiori2024llama3herdmodels}
Aaron Grattafiori, Abhimanyu Dubey, Abhinav Jauhri, Abhinav Pandey, et~al.
\newblock The {Llama} 3 herd of models.
\newblock \emph{arXiv preprint arXiv:2407.21783}, 2024.

\bibitem[OpenAI(2023)]{openai2023gpt}
OpenAI.
\newblock Gpt-4 technical report, 2023.

\end{thebibliography}

\newpage

\appendix

\onecolumn
\begin{center}
{\bf {\LARGE Appendix}}
\end{center}

\section{Additional Experiments}
\label{appx:additional_experiment}

\begin{table*}[ht]
\centering \small{%
\begin{tabular}{ccccccc|cccccc}
\toprule
 &
  \multicolumn{6}{c|}{ImageNet} &
  \multicolumn{6}{c}{ImageNet} \\ \cmidrule(l){2-13} 
 &
  \multicolumn{6}{c|}{w/o ensemble} &
  \multicolumn{6}{c}{w/ ensemble} \\ \cmidrule(l){2-13} 
 Methods&
  \multicolumn{1}{c|}{IN} &
  IN-S &
  IN-R &
  IN-A &
  \multicolumn{1}{c|}{IN-V2} &
  Avg. &
  \multicolumn{1}{c|}{IN} &
  IN-S &
  IN-R &
  IN-A &
  \multicolumn{1}{c|}{IN-V2} &
  Avg. \\ \midrule
Zeroshot &
  \multicolumn{1}{c|}{68.3} &
  \textbf{77.7} &
  50.0 &
  {48.3} &
  \multicolumn{1}{c|}{61.9} &
  59.5 &
  \multicolumn{1}{c|}{-} &
   -&
   -&
   -&
  \multicolumn{1}{c|}{-} &
  - \\
FT &
  \multicolumn{1}{c|}{81.3} &
  70.8 &
  46.4 &
  {46.9} &
  \multicolumn{1}{c|}{71.2} &
  58.8 &
  \multicolumn{1}{c|}{81.8} &
  75.7 &
  49.8 &
  {51.9} &
  \multicolumn{1}{c|}{72.5} &
  62.5 \\
FLYP &
  \multicolumn{1}{c|}{82.6} &
  71.4 &
  48.5 &
  {49.8} &
  \multicolumn{1}{c|}{72.7} &
  60.6 &
  \multicolumn{1}{c|}{82.9} &
  74.1 &
  51.6 &
  {51.3} &
  \multicolumn{1}{c|}{73.6} &
  62.6 \\
CAR-FT &
  \multicolumn{1}{c|}{81.9} &
  75.6 &
  50.5 &
  {51.5} &
  \multicolumn{1}{c|}{72.8} &
  62.5 &
  \multicolumn{1}{c|}{82.0} &
  \underline{77.2} &
  \underline{52.0} &
 {\underline{52.5}} &
   \multicolumn{1}{c|}{72.8} &
  \underline{63.6} \\
CaRot &
  \multicolumn{1}{c|}{\textbf{83.1}} &
  \underline{76.2} &
  \underline{51.3} &
 {\underline{51.9}} &
   \multicolumn{1}{c|}{\textbf{74.3}} &
  \underline{63.4} &
  \multicolumn{1}{c|}{\textbf{83.1}} &
  76.2 &
  51.3 &
  {51.9} &
  \multicolumn{1}{c|}{\textbf{74.3}} &

  63.4 \\ \midrule
\textbf{\sname~(Ours)} &
  \multicolumn{1}{c|}{\underline{82.9}} &
  \textbf{77.7} &
  \textbf{53.7} &
 {\textbf{52.5}} &
   \multicolumn{1}{c|}{\underline{73.8}} &
  \textbf{64.4} &
  \multicolumn{1}{c|}{\underline{83.0}} &
  \textbf{78.3} &
  \textbf{53.0} &
  {\textbf{53.0}} &
  \multicolumn{1}{c|}{\underline{73.8}} &
  \textbf{64.5} \\ \bottomrule
\end{tabular}%
\caption{
\textbf{Weight ensemble results on ImageNet fine-tuned CLIP ViT-B/16.}
We report the results based on the ID validation accuracy. }
\label{tab:ensemble}
}
\end{table*}

\begin{table*}[ht]
\begin{minipage}[t]{0.49\textwidth}
\centering\small
\setlength\tabcolsep{2.5pt} 
\begin{tabular}{ccc}
\toprule
      &  \multicolumn{2}{c}{iWILD}  \\ \cmidrule(l){2-3} 
Mask & ID   & OOD \\ \midrule
    & 49.82& 34.98  \\
\cmark& 50.09 & 37.07 \\ \bottomrule
\end{tabular}%
\vphantom{%
  \begin{tabular}{ccc}
  \toprule
      & \multicolumn{2}{c}{ImageNet} \\ \cmidrule(l){2-3} 
  $\lambda$ & ID & OOD Avg. \\ \midrule
  0.1 & 83.02 & 63.07 \\
  0.5 & 82.88 & 64.41 \\
  1.0 & 82.37 & 64.87 \\ \bottomrule
  \end{tabular}
}
\caption{
\textbf{Effect of positive pair masking.} 
We show the effect of masking positive pairs on WILDS-iWILDCam dataset where zero-shot model performs poorly.}\label{tab:appx_ablation}
\end{minipage}
\hfill
\begin{minipage}[t]{0.49\textwidth}
\centering\small
\setlength\tabcolsep{2.5pt} 
\begin{tabular}{ccc}
\toprule
    & \multicolumn{2}{c}{ImageNet} \\ \cmidrule(l){2-3} 
$\lambda$    & ID          & OOD Avg.       \\ \midrule
0.1 & 83.02       & 63.07          \\
0.5 & 82.88       & 64.41          \\
1.0   & 82.37       & 64.87          \\ \bottomrule
\end{tabular}%
\caption{
\textbf{Ablation on regularization ratio.} 
We sweep on $\lambda \in \{0.1, 0.5, 1.0\}$ to see the effect of our regularization ratio to ID and OOD accuracies on ImageNet.}
\label{tab:appx_ablation_lambda}
\end{minipage}
\end{table*}

\begin{table*}[ht]
\begin{minipage}[t]{0.49\textwidth}
\centering\small
\setlength\tabcolsep{2.5pt} 
\begin{tabular}{lcc|cc}\toprule
&  \multicolumn{2}{c|}{iWILD} &  \multicolumn{2}{c}{FMoW} \\ \cmidrule(l){2-5}
Method       & ID    & OOD   & ID    & OOD  \\ \midrule
Zeroshot  & 8.70  & 11.00 & 20.40 & 18.70 \\
FT        & 47.22 & 35.56 & \textbf{68.64} & 40.15 \\
FLYP      & 48.52 & 36.61 & 68.56 & 40.07 \\
CAR-FT    & 45.75 & 37.02 & 68.44 & 40.69 \\
CaRot &  40.61 &  29.16 &  51.88 &  26.80\\ \midrule
\sname~(ours) &\textbf{50.09}& \textbf{37.07}& 68.41& \textbf{40.96}\\ \bottomrule
\end{tabular}%

\caption{
\textbf{Evaluation on additional domain shifts.} 
Additional results on WILDS-iWILDCam and WILDS FMoW.
}
\label{tab:additional_domain_shift}
\end{minipage}
\hfill
\begin{minipage}[t]{0.49\textwidth}
\centering\small
\setlength\tabcolsep{2.5pt} 
\begin{tabular}{lc|cccc|c}
\toprule
ViT-B/16 & \multicolumn{1}{c|}{IN}& IN-V2 & IN-R & IN-A  & \multicolumn{1}{c|}{IN-S} & Avg. \\ \midrule
Zeroshot  & 68.3 & 61.9 & \underline{77.7} & 50.0 & 48.3 & 59.5\\
FLYP   & 82.6& 72.7 &  71.4 & 48.5 & 49.8   & 60.6 \\ \midrule
{\textbf{StarFT}}: \\
~~~GPT-3.5  &  \textbf{82.9} & \textbf{73.8} & \underline{77.7} & \underline{53.7} & \underline{52.5}   & \underline{64.4} \\
~~~Llama-3.3 & \underline{82.7}& \underline{73.4} & \textbf{78.1} & \textbf{53.9} & \textbf{52.7} & \textbf{64.5} \\
\bottomrule
\end{tabular}
\vphantom{
\begin{tabular}{lcc|cc}\toprule
&  \multicolumn{2}{c|}{iWILD} &  \multicolumn{2}{c}{FMoW} \\ \cmidrule(l){2-5}
Method       & ID    & OOD   & ID    & OOD  \\ \midrule
Zeroshot  & 8.70  & 11.00 & 20.40 & 18.70 \\
FT        & 47.22 & 35.56 & \textbf{68.64} & 40.15 \\
FLYP      & 48.52 & 36.61 & 68.56 & 40.07 \\
CAR-FT    & 45.75 & 37.02 & 68.44 & 40.69 \\
CaRot &  40.61 &  29.16 &  51.88 &  26.80\\ \midrule
\sname~(ours) &\textbf{50.09}& \textbf{37.07}& 68.41& \textbf{40.96}\\ \bottomrule
\end{tabular}%
}
\caption{
\textbf{Ablation on different LM.} 
Additional results using LLaMA-3.3-70B-Instruct for the concept bank generation.}
\label{tab:llm_ablation}
\end{minipage}
\end{table*}

\subsection{Weight ensembling on StarFT}
Weight ensembling of fine-tuned models and zero-shot models is a very simple and straightforward method to boost both ID and OOD accuracies.
We sweep on different choices of $\alpha \in \{0.0, 0.1, 0.2, 0.3, 0.4, 0.5, 0.6, 0.7, 0.8, 0.9, 1.0\}$ and report the result of weight ensembling selected by the best ID accuracy.
As shown in the Table~\ref{tab:ensemble}, ours show best OOD average accuracy among the baselines.
Note that ensembling weight does not improve the ID accuracies of CaRoT and StarFT, as both methods distills beneficial knowledge from zero-shot models that possibly functions like weight ensembling.

\subsection{Ablation on spuriosity regularization ratio}
We study the effect of our regularization $\lambda$ on ID and OOD accuracy. 
Specifically, we sweep on different choices of $\lambda \in \{0.1, 0.5, 1.0\}$.
As expected, if we strengthen our regularization, it improves the OOD accuracy at the cost of ID accuracy.

\subsection{Ablation on positive pair mask on WILDS-iWILDCam}
As shown in Table~\ref{tab:transfer_learning}, the base zero-shot performs poorly on WILDS iWILDCam dataset with ID accuracy of 8.70\%.
Therefore, distilling the poor-performing zero-shot models' high confidences of true class can negatively affect the fine-tuning procedure. Thus, we mask out logits from the true class (\ie positive pairs) to address such issue. As shown in Table~\ref{tab:appx_ablation}, it shows significant performance drop in OOD accuracy compared to the results with masking.

\subsection{Additional domain shifts}
We train \sname on WILDS-iWILDCam~\citep{koh2021wilds, beery2020iwildcam} and WILDS-FMoW~\citep{koh2021wilds, christie2018functional}. 
WILDS-iWILDCam consists of animal images of 182 categories. 
The ID and OOD datasets are collected from different camera traps, leading to the differences in background, illumination, etc.  
WILDS-FMoW consists of remote sensing satellite images of 62 categories regarding different locations. 
The ID and OOD datasets are collected from different geographical sites and time span.
\sname exhibits improved ID and OOD accuracy in iWILDCam and FMoW, showing its efficacy on domain distribution shifts.

\subsection{Different language model}
We conduct an additional ablation study using a more recent LM, LLaMA-3.3-70B-Instruct~\citep{grattafiori2024llama3herdmodels}, in the place of GPT-3.5 for the concept bank generation process of StarFT.
As shown in Table~\ref{tab:llm_ablation}, we observe that spurious descriptions generated by LLaMA-3.3 lead to slightly improved OOD robustness compared to those from GPT-3.5. Nevertheless, StarFT does not exhibit significant degradation even with GPT-3.5, suggesting that it remains robust as long as the language model has a sufficient grasp of spuriosity.
\clearpage
\section{Experimental Details}\label{appx:experimental_detail}
\subsection{Dataset details}
\begin{itemize}[leftmargin=5.5mm]
    \item \textbf{ImageNet}~\citep{russakovsky2015imagenet} is a large scale image dataset where the goal is to classify image into one of 1,000 categories. The training set comprises more than million samples, and the validation set comprises 50,000 images (50 images for each class).
    \item \textbf{ImageNetV2}~\citep{recht2019imagenet} is a newly curated text split of ImageNet which comprises the same 1,000 classes as ImageNet.
    \item
    \textbf{ImageNet-R}~\citep{hendrycks2021faces} is a rendition of ImageNet containing art, cartoons, devianart, graffiti, embroidery, and etc. ImageNet-R has 200 classes of ImageNet.
    \item
    \textbf{ImageNet-A}~\citep{hendrycks2021natural} collects the samples where ResNet make wrong predictions. ImageNet-A has 200 classes of ImageNet.
    \item
    \textbf{ImageNet-Sketch}~\citep{wang2019learning} contains a sketched version of ImageNet. ImageNet-Sketch has the same 1,000 classe as ImageNet.
    \item \textbf{Waterbirds}~\citep{sagawa2020distributionally} contains 2 different categories of birds: waterbird and landbird. Each image contain two different backgrounds, \ie water background and land background. Each background serves as a distinct group for group shift evaluation.
    \item \textbf{CIFAR-10.02}~\citep{zhang2022contrastive} is a combined dataset from two sources: CIFAR-10~\citep{krizhevsky2009learning} and CIFAR-10.2~\citep{lu2020harder}. Each category contains images from both source datasets, and the goal is to classify images into one of 10 categories. Each source is treated as a distinct group for group shift evaluation.
    \item \textbf{PACS}~\citep{li2017deeper} is an image dataset containing 4 different groups: art, painting, cartoon, and sketch. Each group consists of seven categories such as ``guitar'' and ``elephant'' and group accuracies are measured for group shift evaluation.
    \item \textbf{CIFAR-10}~\citep{krizhevsky2009learning} is an image dataset containing 10 different categories. The goal is to classify image into one of 10 categories, such as ``airplane,'' and ``dog.''
    \item \textbf{CIFAR-100}~\citep{krizhevsky2009learning} is an image dataset containing 100 different categories, which is a more challenging dataset compared to CIFAR-10. The goal is to classify image into one of 100 categories, such as ``airplane,'' and ``dog.''
    \item \textbf{Caltech101}~\citep{li_andreeto_ranzato_perona_2022} contains images of various objects such as ``crab'' or ''mandolin''. The goal is to classify image into one of 101 categories.
    \item \textbf{STL10}~\citep{coates2011stl} is an image dataset containing 10 different categories like ``airplane'', where the goal is to classify image into one of 10 categories.
    \item \textbf{StanfordCars}~\citep{krause2013stanfordcars} contains images of cars with different make, model, and year (\eg 2012 Tesla Model S). The goal is to classify the image into one of 196 categories.  
    \item \textbf{Flowers102}~\citep{nilsback2008automated} consists of images of different flowers such as ``hibiscus'' where the goal is to classify into one of 102 categories.
    \item \textbf{WILDS-iWILDCam}~\citep{beery2020iwildcam, koh2021wilds} consists of 182 different categories of animal images. To evaluate distribution shift, we split the dataset into ID and OOD based on the camera used and factors like illumination. Models are fine-tuned with only ID dataset and evaluated for OOD datasets.
    \item \textbf{WILDS-FMoW}~\citep{christie2018functional, koh2021wilds} consists of remote sensing satellite images. We classify satellite images into on of 62 categories, such as ``airport'' or ``zoo''. To evaluate distribution shift, we split the dataset into ID and OOD based on the location and time of their collection. Models are fine-tuned with only ID dataset and evaluated for OOD datasets.
\end{itemize}

\clearpage

\subsection{Training details}
We closely follow the training protocols of \citet{goyal2022finetune} and \citet{wortsman2022robust}. All methods, including StarFT, use an AdamW optimizer and cosine learning rate scheduler. A batch size of 512 and regularization ratio $\lambda$ of 0.5 are applied for ImageNet, while a batch size of 256 and a regularization ratio $\lambda$ of 0.1 is used for all other datasets. We perform an early stopping based on in-distribution (ID) validation accuracy for all the datasets except ImageNet. For ImageNet, we fine-tune the models for 10 epochs and use them for evaluation. 
All experiments were run on a machine with 8 NVIDIA V100 GPUs. For fine-tuning CLIP ViT-B/16 on ImageNet, it took approximately 10 hours to run 10 epochs. We list the hyperparameters for each dataset in Table~\ref{tab:hyperparmeters}.

\begin{table*}[t]
\centering \small
\begin{tabular}{lcccc}
\toprule
Dataset &   Max Epochs  &   Learning Rate   &   Weight Decay    &   Batch Size \\
\midrule
Caltech101  &   100 &   1e-5    &   0.0     &   256 \\
Cars        &   100 &   1e-5    &   0.0     &   256 \\
Flowers     &   20  &   1e-5    &   0.1     &   256 \\
ImageNet    &   10  &   1e-5    &   0.1     &   256 \\
iWILD       &   20  &   1e-5    &   0.2     &   256 \\
FMoW        &   20  &   1e-5    &   0.2     &   256 \\
\bottomrule
\end{tabular}\caption{\textbf{Fine-tuning hyperparameters.}}
\label{tab:hyperparmeters}
\end{table*}

\subsection{Baseline methods}
\begin{itemize}[leftmargin=*]
    \item \textbf{Zero-shot (ZS)}~\citep{radford2021learning}: Zero-shot classifier is constructed by encoding text descriptions of each class using pretrained CLIP text encoder.
    \item \textbf{Standard Fine-tuning (FT)}~\citep{wortsman2022robust}: Starting from the zero-shot linear classification head obtained from pretrained CLIP text encoder, image encoder and linear classifier parameters are fine-tuned with cross-entropy loss. 
    \item \textbf{CAR-FT}~\citep{mao2022contextaware} : CAR-FT regularizes the model during fine-tuning to capture context information alongside the cross-entropy loss. For each dataset, we use prompt templates suggested in FLYP~\citep{goyal2022finetune} as context prompts, which contains dataset specific information .
    \item \textbf{FLYP}~\citep{goyal2022finetune} : Both image encoder and text encoder are trained with contrastive loss used to pretrain CLIP~\citep{radford2021learning}. In training phase, we use selected template (\eg ``\texttt{a photo of a [class]}'') rather than averaging multiple text description templates.
    \item \textbf{CaRot}~\citep{oh2024calibrated} : CaRoT leverages input-dependant label smoothing obtained by self-distillation using an exponential moving average (EMA) teacher model. We set the hyperparmeters regarding EMA update following \citet{oh2024calibrated}.
\end{itemize}

\subsection{Prompt templates}
Table~\ref{tab:appx_templates} provides the examples of templates for gorup shift dataset. For all other dataset, we follow the templates from OpenAI\footnote{
\url{https://github.com/openai/CLIP/blob/main/notebooks/Prompt_Engineering_for_ImageNet.ipynb}}~\citep{radford2021learning}.
\vspace{0.6in}
\begin{table*}[t]
\renewcommand*{\arraystretch}{1.2}
\centering\small
\begin{tabular}{ll}
\toprule
Dataset & Templates \\ \midrule
Waterbirds & ``This is a picture of a \texttt{[class]}.''\\ \midrule
CIFAR-10.02 & ``a \texttt{[class]}.''\\ \midrule
PACS & ``a \texttt{[class]}.''\\
\bottomrule
\end{tabular}\caption{
\textbf{Prompt design for zero-shot classification.} 
We use the following prompt design for group shift dataset following \citet{zhang2022contrastive}.}
\label{tab:appx_templates}
\end{table*}

\clearpage

\section{Utilization of Language Models}\label{appx:language_model}
\subsection{Prompting language models}
We break down our spurious textual description generations into 3 parts. We first get high-level spurious concepts, then we obtain low-level spurious words from these concepts. Finally, we ask for LMs to generate prompt-like spurious descriptions.
Specifically, for high-level spurious concept generation, we prompt LMs with:

\begin{quote}
\begin{footnotesize}
\begin{verbatim}
### Your Task ### You are an image classifier classifying a given natural image into a 
certain class.

Question: List possible spurious correlations while classifying natural images. Answer 
in a word.

Answer: 1.
-
\end{verbatim}
\end{footnotesize}
\end{quote}

After we obtain the spurious concept, we ask for more fine-grained spurious words:
\begin{quote}
\begin{footnotesize}
\begin{verbatim}
### Your Task ### You are an image classifier classifying a given natural image into a
certain class.

Question: List 20 [spurious concept] bias keywords that may lead to a spurious correlation.

Answer: 1.
-
\end{verbatim}
\end{footnotesize}
\end{quote}

We prompt GPT-3.5~\citep{openai2023gpt} 3 runs for each spurious concept, then remove the duplicated spurious words. Finally, we obtain prompt-like spurious textual descriptions by querying:

\begin{quote}
\begin{footnotesize}
\begin{verbatim}
### Your Task ### You design a proper prompt with a given keyword. The base prompt is "a 
photo of a [class]." You append an appropriate postfix to the base prompt with the given 
keyword. You use only one keyword to design each prompt.

Question: Design a prompt with following keywords: [keyword1], [keyword2], ...

Answer: 1.
-
\end{verbatim}
\end{footnotesize}
\end{quote}

\subsection{Spurious textual descriptions}\label{appx:spurious_list}
Table~\ref{tab:appx_spurious_descriptions} provides concrete examples of spurious textual descriptions we obtained via LMs. Specifically, we mainly consider three types of possible spurious concept: ``background'', ``texture'' and ``resolution.'' Each spurious description is appended as a postfix (or prefix) to the base prompt.

\begin{table*}[t]
\renewcommand*{\arraystretch}{1.3}
\centering\small
\begin{tabular}{l|l||l|l}
\toprule
Concept & Spurious descriptions & Concept & Spurious descriptions \\ \midrule
\multirow{30}{*}{Background} &  ``\texttt{[class]} surrounded by flowers''   & \multirow{15}{*}{Texture}   &    ``\texttt{[class]} with a rough texture'' \\
& ``\texttt{[class]} in urban settings'' &  & ``\texttt{[class]} with a smooth surface'' \\
& ``\texttt{[class]} surrounded by greenery'' & & ``\texttt{[class]} feeling soft to the touch'' \\
& ``\texttt{[class]} in snowy landscapes'' & & ``\texttt{[class]} that is hard'' \\
& ``\texttt{[class]} in a rainforest'' & & ``\texttt{[class]} with a fuzzy texture'' \\
& ``\texttt{[class]} on busy roads and highways'' & & ``\texttt{[class]} with prickly features'' \\
& ``\texttt{[class]} in the mountains'' & & ``\texttt{[class]} with a bumpy surface'' \\
& ``\texttt{[class]} in parks and gardens'' & & ``\texttt{[class]} appearing fluffy''\\
& ``\texttt{[class]} in front of historical landmarks'' & & ``\texttt{[class]} with silky textures'' \\
& ``\texttt{[class]} in fields'' & & ``\texttt{[class]} shining brightly''  \\
& ``\texttt{[class]} in the forest and woods'' & & ``\texttt{[class]} with a matte finish'' \\
& ``\texttt{[class]} in the snow'' & & ``\texttt{[class]} with wavy patterns''  \\
& ``\texttt{[class]} on the beach'' & & ``\texttt{[class]} with knotty features'' \\
& ``\texttt{[class]} in a farmland'' & & ``\texttt{[class]} with lumpy textures''\\
& ``\texttt{[class]} on the sand and beach'' & & ``\texttt{[class]} with a sleek appearance'' \\ \cline{3-4}
& ``\texttt{[class]} with animals in the frame'' & \multirow{15}{*}{Resolution} & ``blurred \texttt{[class]}'' \\ 
& ``\texttt{[class]} in entertainment centers'' & & ``blurry \texttt{[class]}''\\
& ``\texttt{[class]} in a suburban area'' & & ``\texttt{[class]} with brightness'' \\
& ``\texttt{[class]} under a blue sky'' & & ``dark \texttt{[class]}'' \\
& ``\texttt{[class]} surrounded by butterflies'' & & ``distorted \texttt{[class]}'' \\
& ``\texttt{[class]} with clouds in the background'' & & ``high resolution of \texttt{[class]}'' \\
& ``\texttt{[class]} in cloudy weather'' & & ``low resolution of \texttt{[class]}'' \\
& ``\texttt{[class]} in the desert'' & & ``overexposed \texttt{[class]}'' \\
& ``\texttt{[class]} among rocks'' & & ``underexposed \texttt{[class]}'' \\
& ``\texttt{[class]} in a forest'' & & ``\texttt{[class]} with shadow'' \\
& ``\texttt{[class]} in an industrial setting'' & & ``\texttt{[class]} with low visibility'' \\
& ``\texttt{[class]} in residential areas'' & & ``\texttt{[class]} with inconsistent lighting'' \\
& ``\texttt{[class]} in sports stadiums and arenas'' & & ``cropped \texttt{[class]}'' \\
& ``\texttt{[class]} during sunset'' & & ``\texttt{[class]} with cropping'' \\
& ``\texttt{[class]} surrounded by trees'' & & ``monochrome \texttt{[class]}'' \\ \bottomrule
\end{tabular}\caption{
\textbf{List of spurious descriptions.} We provide the examples of spurious descriptions obtained using LMs, categorized by spurious concepts (\ie background, texture, and resolution).}
\label{tab:appx_spurious_descriptions}
\end{table*}

\section{Discussions}
\label{appx:limitations}

\paragraph{Limitations. } Since \sname tackles general spurious concepts based on the textual descriptions, further investigation into the construction of these spurious concepts is needed. 
We acknowledge that the hallucination issues in LLMs may affect the effectiveness of our approach. We nevertheless note that our method design can be inherently robust to such issues, given that our objective only regularizes fine-tuning to “not to learn” from the spuriosity textual embeddings (identified by LLMs): it is still possible to learn from other embeddings even when there are some hallucinated descriptors in the prompt set. This intuition is supported by the consistent performance gains from our experimental validation. However, a deeper assessment of the quality of spurious information would be a promising direction to further boost the effectiveness. We expect that future work can discover and provide guidelines acquiring (or filtering) spurious concepts for specific downstream tasks based on our insights. 
Also, extending our work to various other of VLMs (\eg ALIGN~\citep{jia2021scaling}) beyond CLIP would be an interesting future work.

\paragraph{Broader impact. }  We argue that understanding capabilities and limitations of foundation models (\eg CLIP), especially when fine-tuned for specific downstream tasks, is needed. Due to the promising transfer to downstream tasks, individuals may get the sense that these pre-trained models can be successfully applied to any desired downstream tasks. However, our work observes that spuriosity often arises during fine-tuning and highlights the importance of preventing the model to learn undesirable features. Further discussions within this context is needed for better understanding and potential uses of foundation models.
\label{appx:broader_impact}

\end{document}